\documentclass[acmsmall]{acmart}

\newcommand{\chapquote}[3]{\begin{quotation} \textit{#1} \end{quotation}}
\usepackage{setspace}
\AtBeginDocument{%
  \providecommand\BibTeX{{%
    \normalfont B\kern-0.5em{\scshape i\kern-0.25em b}\kern-0.8em\TeX}}}
\usepackage{enumitem}
\usepackage{mathtools}
\usepackage{bbm}

\begin{document}



\title{Event Prediction in the Big Data Era: A Systematic Survey}

\author{Liang Zhao}

\email{liang.zhao@emory.edu}
\affiliation{%
  \institution{Emory University}
}









\begin{abstract}
Events are occurrences in specific locations, time, and semantics that nontrivially impact either our society or the nature, such as earthquakes, civil unrest,  system failures, pandemics, and crimes. It is highly desirable to be able to anticipate the occurrence of such events in advance in order to reduce the potential social upheaval
and damage caused. Event prediction, which has traditionally been prohibitively challenging, is now becoming a viable option in the big data era and is thus experiencing rapid growth, also thanks to advances in high performance computers and new Artificial Intelligence techniques. There is a large amount of existing work that focuses on addressing the challenges involved, including heterogeneous multi-faceted outputs, complex (e.g., spatial, temporal, and semantic) dependencies, and streaming data feeds. Due to the strong interdisciplinary nature of event prediction problems, most existing event prediction methods were initially designed to deal with specific application domains, though the techniques and evaluation procedures utilized are usually generalizable across different domains. However, it is imperative yet difficult to cross-reference the techniques across different domains, given the absence of a comprehensive literature survey for event prediction. This paper aims to provide a systematic and comprehensive survey of the technologies, applications, and evaluations of event prediction in the big data era. First, systematic categorization and summary of existing techniques are presented, which facilitate domain experts' searches for suitable techniques and help model developers consolidate their research at the frontiers. Then, comprehensive categorization and summary of major application domains are provided to introduce wider applications to model developers to help them expand the impacts of their research. Evaluation metrics and procedures are summarized and standardized to unify the understanding of model performance among stakeholders, model developers, and domain experts in various application domains. Finally, open problems and future directions for this promising and important domain are elucidated and discussed.


\end{abstract}




\maketitle

 \section{Introduction}
An event is a real-world occurrence that takes place in a specific location and time that relates to a particular topic. Events can range from large-scale (e.g., civil unrest events and earthquakes), to medium-scale (e.g., system failures and crime incidents), to small-scale (e.g., authentication events and individual actions) occurrences ~\cite{allan1998line}. Event analytics are important in domains as different as healthcare, business, cybersphere, politics, and entertainment, influencing almost every corner of our lives~\cite{yamaguchi1991event}. The analysis of events has thus been attracting huge attention over the last few decades and can be categorized in terms of their timeliness for various research directions, such as event summarization, detection, and prediction. Unlike retrospective analyses such as event summarization and detection~\cite{chakrabarti2011event}, event prediction focuses on anticipating events in the future and is the focus of this survey. Accurate anticipation of future events enables one to maximize the benefits and minimize the losses associated with some event in the future, bringing huge benefits for both society as a whole and individual members of society in key domains such as disease prevention~\cite{qiao2018pairwise}, disaster management~\cite{mallouhy2019major}, business intelligence~\cite{yang2019using}, and economics stability~\cite{bialonski2015data}.


\chapquote{``Prediction is very difficult, especially if it's about the future.'' \ \ \         -- Niels Bohr, 1970}{}{}

Event prediction has traditionally been prohibitively challenging across different domains, due to the lack or incompleteness of our knowledge regarding the true causes and mechanisms driving event occurrences in most domains. With the advent of the big data era, however, we now enjoy unprecedented opportunities that open up many alternative approaches for dealing with event prediction problems, sidestepping the need to develop a complete understanding of the underlying mechanisms of event occurrence. Based on large amounts of data on historical events and their potential precursors, event prediction methods typically strive to apply predictive mapping to build on these observations to predict future events, utilizing predictive analysis techniques from domains such as machine learning, data mining, pattern recognition, statistics, and other computational models~\cite{bishop2006pattern,han2011data,baeza1999modern}. Event prediction is currently experiencing extremely rapid growth, thanks to advances in sensing techniques (physical sensors and social sensors), prediction techniques (Artificial Intelligence, especially Machine Learning), and high performance computing hardware~\cite{flouris2017issues}.


Event prediction in big data is a difficult problem that requires the invention and integration of related techniques to address the serious challenges caused by its unique characteristics, including: \textbf{1) Heterogeneous multi-output predictions.} Event prediction methods usually need to predict multiple facets of events including their time, location, topic, intensity, and duration, each of which may utilize a different data structure~\cite{ramakrishnan2014beating}. This creates unique challenges, including how to jointly predict these heterogeneous yet correlated facets of outputs. Due to the rich information in the outputs, label preparation is usually a highly labor-intensive task performed by human annotators, with automatic methods introducing numerous errors in items such as event coding. So, how can we improve the label quality as well as the model robustness under corrupted labels? The multi-faceted nature of events make event prediction a multi-objective problem, which raises the question of how to properly unify the prediction performance on different facets. It is also challenging to verify whether a predicted event ``matches'' a real event, given that the various facets are seldom, if ever, 100\% accurately predicted. So, how can we set up the criteria needed to discriminate between a correct prediction (``true positive'') and a wrong one (``false positive'')? \textbf{2) Complex dependencies among the prediction outputs.} Beyond conventional isolated tasks in machine learning and predictive analysis, in event prediction the predicted events can correlate to and influence each other~\cite{matsubara2012fast}. For example, an ongoing traffic incident event could cause congestion on the current road segment in the first 5 minutes but then lead to congestion on other contiguous road segments 10 minutes later. Global climate data might indicate a drought in one location, which could then cause famine in the area and lead to a mass exodus of refugees moving to another location. So, how should we consider the correlations among future events? \textbf{3) Real-time stream of prediction tasks.} Event prediction usually requires continuous monitoring of the observed input data in order to trigger timely alerts of future potential events~\cite{salfner2010survey}. However, during this process the trained prediction model gradually becomes outdated, as real world events continually change dynamically, concepts are fluid and distribution drifts are inevitable. For example, in September 2008 21\% of the United States population were social media users, including 2\% of those over 65. However, by May 2018, 72\% of the United States population were social media users, including 40\% of those over~\cite{center2017social}. Not only the data distribution, but also the number of features and input data sources can also vary in real time. Hence, it is imperative to periodically upgrade the models, which raises further questions concerning how to train models based on non-stationary distributions, while balancing the cost (such as computation cost and data annotation cost) and timeliness?
In addition, event prediction involves many other common yet open challenges, such as imbalanced data (for example data that lacks positive labels in rare event prediction)~\cite{vilalta2002predicting}, data corruption in inputs~\cite{zhao2016hierarchical}, the uncertainty of predictions~\cite{bilovs2019uncertainty}, longer-term predictions (including how to trade-off prediction accuracy and lead time)~\cite{bisset2009epifast}, trade-offs between precision and recall~\cite{ramakrishnan2014beating}, and how to deal with high-dimensionality~\cite{zhao2018distant} and sparse data involving many unrelated features~\cite{wang2018graph}. Event prediction problems provide unique testbeds for jointly handling such challenges.





In recent years, a considerable amount of research has been devoted to event prediction technique development and applications, in order to address the aforementioned challenges~\cite{ning2019spatio}. Recently, there has been a surge of research that both proposes and applies new approaches in numerous domains, though event prediction techniques are generally still in their infancy. Most existing event prediction methods have been designed for a specific application domains, but their approaches are usually general enough to handle problems in other application domains. Unfortunately, it is difficult to cross-reference these techniques across different application domains serving totally different communities. Moreover, the quality of event prediction results require sophisticated and specially-designed evaluation strategies due to the subject matter's unique characteristics, for example its multi-objective nature (e.g., accuracy, resolution, efficiency, and lead time) and heterogeneous prediction results (e.g., heterogeneity and multi-output). As yet, however, we lack a systematic standardization and comprehensive summarization approaches with which to evaluate the various event prediction methodologies that have been proposed. This absence of a systematic summary and taxonomy of existing techniques and applications in event prediction causes major problems for those working in the field who lacks clear information on the existing bottlenecks, traps, open problems, and potentially fruitful future research directions. 




To overcome these hurdles and facilitate the development of better event prediction methodologies and applications, this survey paper aims to provide a comprehensive and systematic review of the current state of the art for event prediction in the big data era. The paper's major contributions include:
\begin{itemize}[leftmargin=*]
    \item \textbf{A systematic categorization and summarization of existing techniques.} Existing event prediction methods are categorized according to their event aspects (time, location, and semantics), problem formulation, and corresponding techniques to create the taxonomy of a generic framework. Relationships, advantages, and disadvantages among different subcategories are discussed, along with details of the techniques under each subcategory. The proposed taxonomy is designed to help domain experts locate the most useful techniques for their targeted problem settings.
    \item \textbf{A comprehensive categorization and summarization of major application domains.} The first taxonomy of event prediction application domains is provided. The practical significance and problem formulation are elucidated for each application domain or subdomain, enabling it to be easily mapped to the proposed technique taxonomy. This will help data scientists and model developers to search for additional application domains and datasets that they can use to evaluate their newly proposed methods, and at the same time expand their advanced techniques to encompass new application domains.
    \item \textbf{Standardized evaluation metrics and procedures.} Due to the nontrivial structure of event prediction outputs, which can contain multiple fields such as time, location, intensity, duration, and topic, this paper proposes a set of standard metrics with which to standardize existing ways to pair predicted events with true events. Then additional metrics are introduced and standardized to evaluate the overall accuracy and quality of the predictions to assess how close the predicted events are to the real ones.
    \item \textbf{An insightful discussion of the current status of research in this area and future trends.} Based on the comprehensive and systematic survey and investigation of existing event prediction techniques and applications presented here, an overall picture and the shape of the current research frontiers are outlined. The paper concludes by presenting fresh insights into the bottleneck, traps, and open problems, as well as a discussion of possible future directions.
\end{itemize}




\subsection{Related Surveys}

This section briefly outlines previous surveys in various domains that have some relevance to  event prediction in big data in three categories, namely: 1. event detection, 2. predictive analytics, and 3. domain-specific event prediction. 

Event detection has been an extensively explored domain with over many years. Its main purpose is to detect historical or ongoing events rather than to predict as yet unseen events in the future~\cite{sakaki2010earthquake,yamaguchi1991event}. Event detection typically focuses on pattern recognition~\cite{bishop2006pattern}, anomaly detection~\cite{han2011data}, and clustering~\cite{han2011data}, which are very different from those in event prediction. There have been several surveys of research in this domain  in the last decade \cite{alevizos2017probabilistic,deng2015overview,michelioudakis2016event,atefeh2015survey}. For example, Deng et al.~\cite{deng2015overview} and Atefeh and Khreich~\cite{atefeh2015survey} provided  overviews of event extraction techniques in social media, while Michelioudakis et al.~\cite{michelioudakis2016event} presented a survey of event recognition with uncerntainty. Alevizos et al.~\cite{alevizos2017probabilistic} provided a comprehensive literature review of event recognition methods using probabilistic methods.

Predictive analysis covers the prediction of target variables given a set of dependent variables. These target variables are typically homogeneous scalar or vector data for describing items  such  as economic indices, housing prices, or sentiments. The target variables may not necessarily be values in the future. Larose~\cite{larose2015data} provides a good tutorial and survey for this domain. Predictive analysis can be broken down into subdomains such as structured prediction~\cite{bishop2006pattern}, spatial prediction~\cite{jiang2018survey}, and sequence prediction~\cite{goodfellow2016deep}, enabling users to handle different types of structure for the target variable. F{\"u}l{\"o}p et al.~\cite{fulop2010survey} provided a survey and categorization of applications that utilize predictive analytics techniques to perform event processing and detection, while Jiang~\cite{jiang2018survey} focused on spatial prediction methods that predict the indices that have spatial dependency. Baklr et al.~\cite{bakir2007predicting} summarized the literature on predicting structural data such as geometric objects and networks, and Arias et al.~\cite{arias2014forecasting} Phillips et al.~\cite{phillips2017using}, and Yu and Kak~\cite{yu2012survey} all proposed the techniques for predictive analysis using social data.

As event prediction methods are typically motivated by specific application domains, there are a number of surveys event predictions for domains such as flood events~\cite{cloke2009ensemble}, social unrest~\cite{blair2020forecasting}, wind power ramp forecasting~\cite{ferreira2011survey}, tornado events~\cite{doswell1993tornado}, temporal events without location information~\cite{gmati2019taxonomy}, online failures~\cite{salfner2010survey}, and business failures~\cite{alaka2016methodological}. However, in spite of its promise and its rapid growth in recent years, the domain of event prediction in big data still suffers from the lack of a comprehensive and systematic literature survey covering all its various aspects, including relevant techniques, applications, evaluations, and open problems. 

\subsection{Outline}
The remainder of this article is organized as follows. Section \ref{sec:formulation} presents generic problem formulations for event prediction and the evaluation of event prediction results. Section \ref{sec:techniques} then presents a taxonomy and comprehensive description of event prediction techniques, after which Section \ref{sec:applications} categorizes and summarizes the various applications of event prediction. Section \ref{sec:discussion} lists the open problems and suggests future research directions and this survey concludes with a brief summary in Section \ref{sec:conclusions}.


 \section{Problem Formulation and Performance Evaluations}
\label{sec:formulation}
This section begins by examining the generic denotation and formulation of the event prediction problem (Section \ref{sec:problem_formulation}) and then considers way to standardize event prediction evaluations (Section \ref{sec:evaluation}). 
\subsection{Problem Formulation}
\label{sec:problem_formulation}
An event refers to a real-world occurrence that happens at some specific time and location with specific semantic topic~\cite{yamaguchi1991event}. We can use $y=(t,l,s)$ to denote an event where its time $t\in\mathcal T$, its location $l\in\mathcal{L}$, and its semantic meaning $s\in\mathcal{S}$. Here, $\mathcal T$, $\mathcal L$, and $\mathcal S$ represent the time domain, location domain, and semantic domain, respectively. Notice that these domains need to have very general meanings that cover a wide range of types of entities. For example, the location $\mathcal L$ can include any features that can be used to locate the place of an event in terms of a point or a neighborhood in either Euclidean space (e.g., coordinate and geospatial region) or non-Euclidean space (e.g., a vertex or subgraph in a network). Similarly, the semantic domain $\mathcal S$ can contain any type of semantic features that are useful when elaborating the semantics of an event's various aspects, including its actors, objects, actions, magnitude, textual descriptions, and other profiling information. For example, \emph{(``11am, June 1, 2019'', ``Hermosillo, Sonora, Mexico'', ``Student Protests'')} and \emph{(``June 1, 2010'', “Berlin, Germany'', ``Red Cross helps pandemics control'')} denote the time, location, and semantics, for two events respectively.

An event prediction system requires inputs that could indicate future events, called event indicators, and these could contain both critical information on events that precede the future event, known as precursors, as well as irrelevant information~\cite{ramakrishnan2014beating,ghil2011extreme}. Event indicator data can be denoted as $X\subseteq\mathcal{T}\times\mathcal{L}\times\mathcal{F}$, where $\mathcal{F}$ is the domain of the features other than location and time. If we denote the current time as $t_{\mathrm{now}}$ and define the past time and future time as $\mathcal{T}^{-}\equiv\{t|t\le t_{\mathrm{now}},t\in\mathcal{T}\}$ and $\mathcal{T}^{+}\equiv\{t|t> t_{\mathrm{now}},t\in\mathcal{T}\}$, respectively, the event prediction problem can now be formulated as follows:
\begin{definition}[Event Prediction] Given the event indicator data $X\subseteq\mathcal{T}^{-}\times\mathcal{L}\times\mathcal{F}$ and historical event data $Y_0\subseteq\mathcal{T}^{-}\times\mathcal{L}\times\mathcal{S}$, event prediction is a process that outputs a set of predicted future events $\hat Y\subseteq\mathcal{T}^{+}\times\mathcal{L}\times\mathcal{S}$, such that for each predicted future event $\hat y=(t,l,s)\in \hat Y$ where $t>t_{\mathrm{now}}$.
\end{definition}
Not every event prediction method necessarily focuses on predicting all three domains of time, location, and semantics simultaneously, but may instead predict any part of them. For example, when predicting a clinical event such as the recurrence of disease in a patient, the event location might not always be meaningful~\cite{qiao2018pairwise}, but when predicting outbreaks of seasonal flu, the semantic meaning is already known and the focus is the location and time~\cite{bisset2009epifast} and when predicting political events, sometimes the location, time, and semantics (e.g., event type, participant population type, and event scale) are all necessary~\cite{ramakrishnan2014beating}. Moreover, due to the intrinsic nature of time, location, and semantic data, the prediction techniques and evaluation metrics of them are necessarily different, as described in the following.

\subsection{Event Prediction Evaluation}
\label{sec:evaluation}
Event Prediction Evaluation essentially investigates the goodness of fit for a set of predicted events $\hat Y$ against real events $Y$. Unlike the outputs of conventional machine learning models such as the simple scalar values used to indicate class types in classification or numerical values in regression, the outputs of event prediction are entities with rich information. Before we evaluate the quality of prediction, we need to first determine the pairs of predictions and the labels that will be used  for the comparison. Hence, we must first optimize the process of matching predictions and real events (Section \ref{sec:matching}) before evaluating the prediction error and accuracy (Section \ref{sec:acc_err}).

\begin{figure}[htb]
  \centering
    \includegraphics[width=0.7\textwidth]{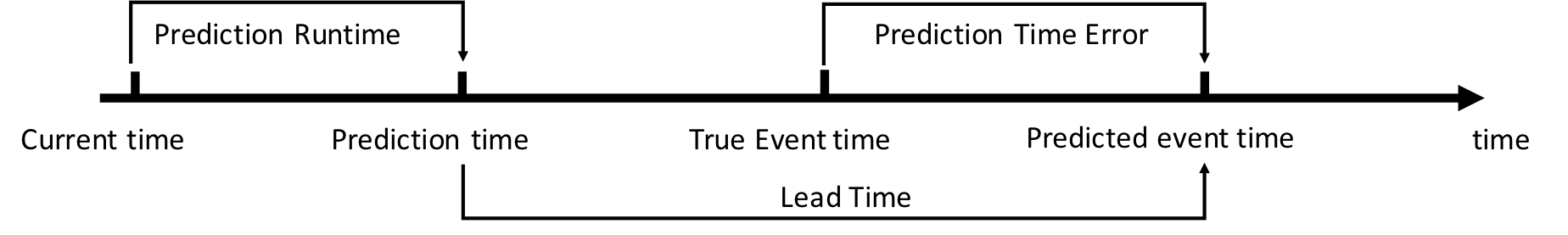}\vspace{-0.3cm}
\caption{Various types of times in event prediction.\vspace{-0.3cm}}
\label{figure:planned_event}
\end{figure}

\subsubsection{Matching predicted events and real events.}
\label{sec:matching}
\begin{figure}[htb]
  \centering
    \includegraphics[width=\textwidth]{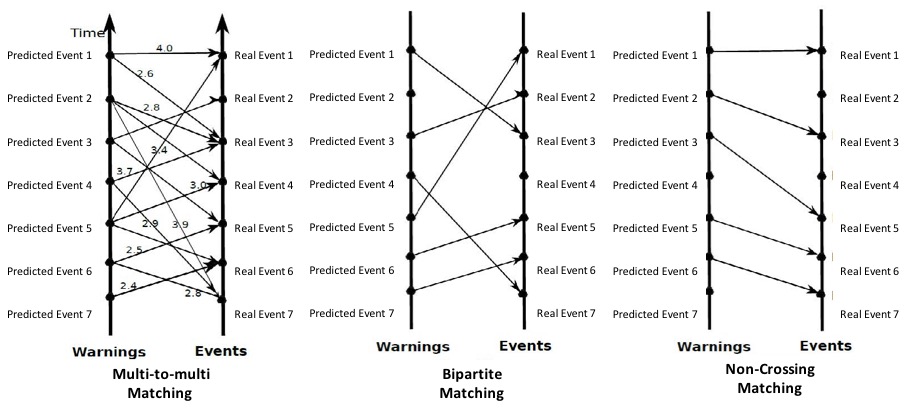}\vspace{-0.3cm}
\caption{Generic framework for hierarchical RNN-based event forecasting.\vspace{-0.3cm}}
\label{fig:event_matching}
\end{figure}

The following two types of matching are typically used: 
\begin{itemize}
\item\textbf{Prefixed matching:} The predicted events will be matched with the corresponding ground-true real events if they share some key attributes. For example, for event prediction at a particular time and location point, we can evaluate the prediction against the ground truth for that time and location. This type of matching is most common when each of the prediction results can be uniquely distinguished along the predefined attributes (for example, location and time) that have a limited number of possible values, so that one-on-one matching between the predicted and real events are easily achieved~\cite{zhao2015multi,abuella2019forecasting}. For example, to evaluate the quality of a predicted event on June 1, 2019 in San Francisco, USA, the true event occurrence on that date in San Francisco can be used for the evaluation.

\item\textbf{Optimized matching:} In situations where one-on-one matching is not easily achieved for any event attribute, then the set of predicted events might need to assess the quality of the match achieved with the set of real events, via an optimized matching strategy~\cite{ramakrishnan2014beating,radinsky2012learning}. For example, consider two predictions, \textbf{Prediction 1:} (``9am, June 4, 2019'', ``Nogales, Sonora, Mexico'', ``Worker Strike''), and \textbf{Prediction 2}: (``11am, June 1, 2019'', ``Hermosillo, Sonora, Mexico'', ``Student Protests''). The two ground truth events that these can usefully be compared with are \textbf{Real Event 1}: (``9am, June 1, 2019'', ``Hermosillo, Sonora, Mexico'', ``Teacher Protests''), and \textbf{Real Event 2}: (``June 4, 2019'', ``Navojoa, Sonora, Mexico'', ``General-population Protest''). None of the predictions are an exact match for any of the attributes of the real events, so we will need to find a ``best'' matching among them, which in this case is between \textbf{Prediction 1} and \textbf{Real Event 2} and \textbf{Prediction 2} and \textbf{Real Event 1}. This type of matching allows some degree of inaccuracy in the matching process by quantifying the distance between the predicted and real events among all the attribute dimensions. The distance metrics are typically either Euclidean distance~\cite{jiang2018survey} or some other distance metric~\cite{han2011data}. Some researchers have hired referees to manually check the similarity of semantic meanings~\cite{radinsky2012learning}, but another way is to use event coding to code the events into an event type taxonomy and then consider a match to have been achieved if the event type matches~\cite{compton2014using}.

Based on the distance between each pair of predicted and real events, the optimal matching will be the one that results in the smallest average distance~\cite{ning2019spatio}. However, suppose there are $m$ predicted events and $n$ real events, then there can be as many as $2^{m\cdot n}$ possible ways of matching, making it prohibitively difficult to find the optimal solution. Moreover, there could be different rules for matching. For example, the ``multiple-to-multiple'' rule shown in Figure \ref{fig:event_matching}(a) allows one predicted (real) event to match multiple real (predicted) events~\cite{radinsky2013mining}, while ``Bipartite matching'' only allows one-to-one matching between predicted and real events (Figure \ref{fig:event_matching}(b)). ``Non-crossing matching'' requires that the real events matched by the predicted events follow the same chronological order (Figure \ref{fig:event_matching}(c)). In order to utilize any of these types of matching, researchers have suggested using event matching optimization to learn the optimal set of ``(real event, predicted event)'' pairs~\cite{muthiah2016embers}.
\end{itemize}

\subsubsection{Metrics of Effectiveness}
\label{sec:acc_err}
The effectiveness of the event predictions is evaluated in terms of two indicators: 1) Goodness of Matching, which evaluates performance metrics such as the number and percentage of matched events~\cite{bishop2006pattern}, and 2) Quality of Matched Predictions, which evaluates how close the predicted event is to the real event for each pair of matched events~\cite{ramakrishnan2014beating}.
\begin{itemize}
\item\textbf{Goodness of Matching.} A \emph{true positive} means a real event has been successfully matched by a predicted event; if a real event has not been matched by any predicted event, then it is called a \emph{false negative} and a \emph{false positive} means a predicted event has failed to match any real event, which is referred to as a \emph{false alarm}. Assume the total number of predictions is $N$, the number of real events is $\hat N$, the number of true positives is $N_{TP}$, the number of false negatives is $N_{FN}$ and the number of false positives is $N_{FP}$. Then, the following key evaluation metrics can be calculated: Prediction=$N_{TP}/(N_{TP}+N_{FP})$, Recall=$N_{TP}/(N_{TP}+N_{FN})$, F-measure = $2\cdot\mathrm{Precision}\cdot\mathrm{Recall}/(\mathrm{Precision}+\mathrm{Recall})$. Other measurements such as the area under the ROC curves are also commonly used~\cite{bishop2006pattern}. This approach can be extended to include other items such as multi-class precision/recall, and Precision/Recall at Top $K$~\cite{zhou2015pattern,jans2012skip,lei2019event,acharya2017causal}.



\item\textbf{Quality of Matched Predictions.}
If a predicted event matches a real one, it is common to go on to evaluate how close they are. This reflects the quality of the matched predictions, in terms of different aspects of the events. Event time is typically a numerical values and hence can be easily measured in terms of metrics such as mean squared error, root mean squared error, and mean absolute error~\cite{bishop2006pattern}. This is also the case for location in Euclidean space, which can be measured in terms of the Euclidean distance between the predicted point (or region) and the real point (or region). Some researchers consider the administrative unit resolution. For example, a predicted location (``New York City'', ``New York State'', ``USA'') has a distance of 2 from the real location (``Los Angeles'', ``California'', ``USA'')~\cite{zhao2016multi}. Others prefer to measure multi-resolution location prediction quality as follows: $(1/3)(l_{country}+l_{country}\cdot l_{state}+l_{country}\cdot l_{state}\cdot l_{city})$, where $l_{city}$, $l_{state}$, and $l_{country}$ can only be either $0$ (i.e., no match to the truth) or $1$ (i.e., completely matches the truth)~\cite{ramakrishnan2014beating}. For a location in non-Euclidean space such as a network~\cite{shao2017efficient}, the quality can be measured in terms of the shortest path length between the predicted node (or subgraph) and the real node (or subgraph), or by the F-measure between the detected subsets of nodes against the real ones, which is similar to the approach for evaluating community detection~\cite{han2011data}. For event topics, in addition to conventional ways of evaluating continuous values such as population size, ordinal values such as event scale, and categorical values such as event type, actors, and actions, as well as more complex semantic values such as texts, can be evaluated using Natural Language Process measurements such as edit distance, BLEU score, Top-$K$ precision, and ROUGE~\cite{allan1998line}.
\end{itemize}

















 \section{Event Prediction Techniques}
\label{sec:techniques}
\begin{figure}[htb]
  \centering
    \includegraphics[width=\textwidth]{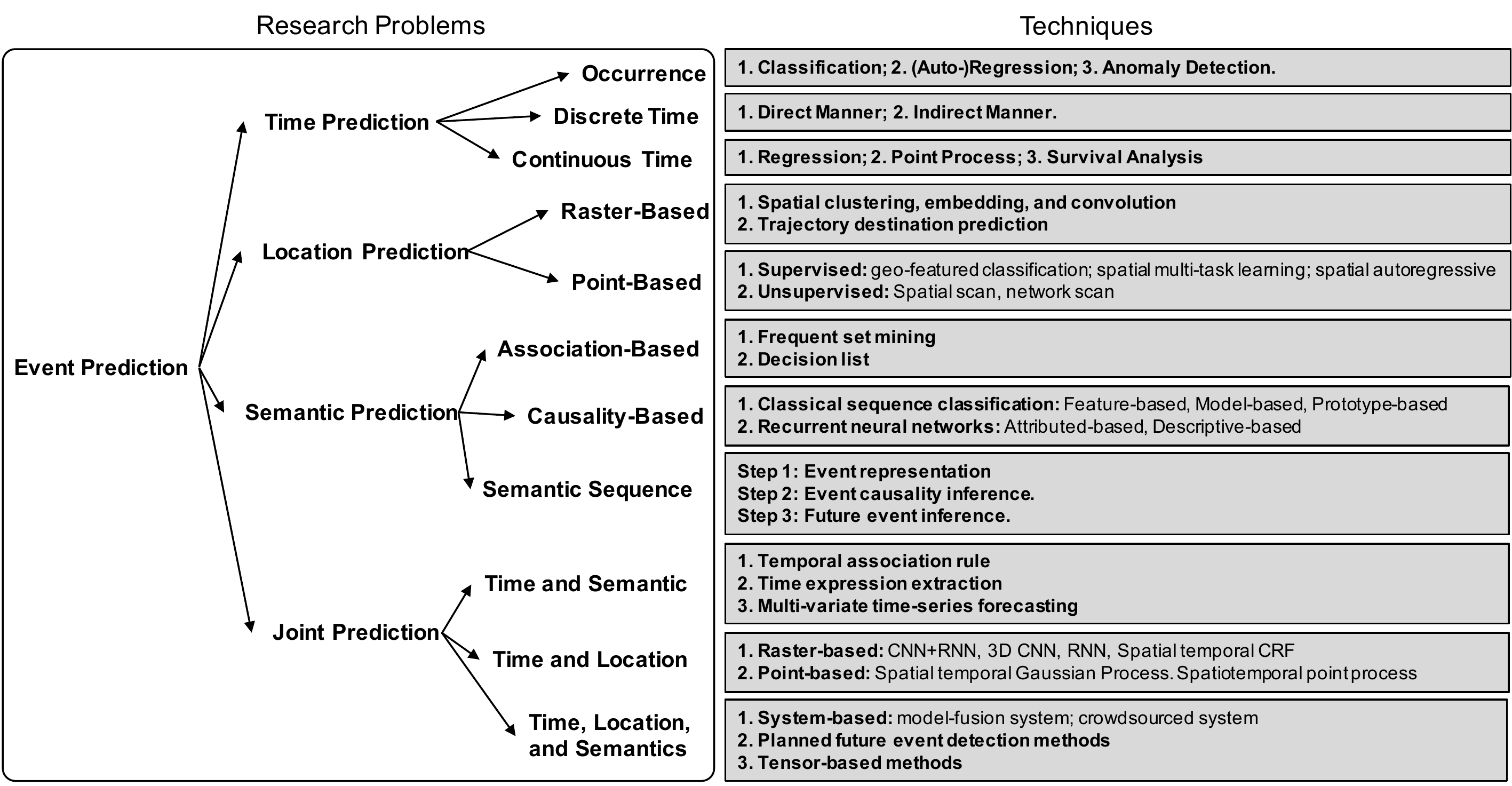}\vspace{-0.3cm}
\caption{Taxonomy of event prediction problems and techniques.\vspace{-0.3cm}}
\label{fig:taxonomy}
\end{figure}

This section focuses on the taxonomy and representative techniques utilized for each category and subcategory. Due to the heterogeneity of the prediction output, the technique types depend on the type of output to be predicted, such as time, location, and semantics. As shown in Figure \ref{fig:taxonomy}, all the event prediction methods are classified in terms of their goals, including time, location, semantics, and the various combinations of these three. These are then further categorized in terms of the output forms of the goals of each and the corresponding techniques normally used, as elaborated in the following.

\subsection{Time Prediction}
Event time prediction focuses on predicting when future events will occur. Based on their time granularity, time prediction methods can be categorized into three types: 1) event Occurrence: Binary-valued prediction on whether an event does or does not occur in a future time period; 2) discrete-time prediction: in which future time slot will the event occur; and 3) continuous-time prediction: at which precise time point will the future event occur.
\subsubsection{Occurrence Prediction} Occurrence prediction is arguably the most extensive, classical, and generally simplest type of event time prediction task~\cite{arias2014forecasting}. It focuses on identifying whether there will be event occurrence (positive class) or not (negative class) in a future time period~\cite{zhao2015multi}. This problem is usually formulated as a binary classification problem, although a handful of other methods instead leverage anomaly detection or regression-based techniques.


\noindent\textbf{1. Binary classification.} Binary classification methods have been extensively explored for event occurrence prediction. The goal here is essentially to estimate and compare the values of $f(y=``Yes''|X)$ and $f(y=``No''|X)$, where the former denotes the score or likelihood of event occurrence given observation $X$ while the latter corresponds to no event occurrence. If the value of the former is larger than the latter, then a future event occurrence is predicted, but if not, there is no event predicted. To implement $f$, the methods typically used rely on discriminative models, where dedicated feature engineering is leveraged to manually extract potential event precursor features to feed into the models. Over the years, researchers have leveraged various binary classification techniques ranging from the simplest threshold-based methods~\cite{zhong2019prediction,laxman2008stream}, to more sophisticated methods such as logistic regression~\cite{zhao2016hierarchical,ballings2012customer}, Support Vector Machines~\cite{inceoglu2018using}, (Convolutional) Neural Networks~\cite{lin2018grid,catling2020temporal}, and decision trees~\cite{de2018new,santiso2014adverse}. In addition to discrminative models, generative models~\cite{zhao2015spatiotemporal,antunes2003bayesian} have also been used to embed human knowledge for classifying event occurrences using Bayesian decision techniques. Specifically, instead of assuming that the input features are independent, prior knowledge can also be directly leveraged to establish Bayesian networks among the observed features and variables based on graphical models such as (semi-)hidden Markov models~\cite{daidone2006hidden,zhao2015spatiotemporal,salfner2007using} and autoregresive logit models~\cite{taylor2017probabilistic}. The joint probabilities $p(y=``Yes'',X)$ of $p(y=``No'',X)$ can thus be estimated using graphical models, and then utilized to estimate $f(y=``Yes''|X)=p(y=``Yes''|X)$ and $f(y=``No''|X)=p(y=``No''|X)$ using Bayesian rules~\cite{bishop2006pattern}.

\noindent\textbf{2. Anomaly detection.} Alternatively, anomaly detection can also be utilized to learn a ``prototype'' of normal samples (typical values corresponding to the situation of no event occurrence), and then identify if any newly-arriving sample is close to or distant from the normal samples, with distant ones being identified as future event occurrences. Such methods are typically utilized to handle ``rare event'' occurrences, especially when the training data is highly imbalanced with little to no data for ``positive'' samples. Anomaly detection techniques such as one-classification~\cite{shin2019autoencoder} and hypotheses testing~\cite{hughes2002improved,shao2017efficient} are often utilized here.

\noindent\textbf{3. Regression.} In addition to simply predicting the occurrence or not, some researchers have sought to extend the binary prediction problem to deal with ordinal and numerical prediction problems, including event count prediction based on (auto)regression~\cite{ertugrul2018forecasting}, event size prediction using linear regression~\cite{yonamine2013predicting}, and event scale prediction using ordinal regression~\cite{gao2018incomplete}.

\subsubsection{Discrete-time Prediction} In many applications, practitioners want to know the approximate time (i.e. the date, week, or month) of future events in addition to just their occurrence. To do this, the time is typically first partitioned into different time slots and the various methods focus on identifying which time slot future events are likely to occur in. Existing research on this problem can be classified into either direct or indirect approaches.

\noindent\textbf{1. Direct Approaches}. These of methods discretize the future time into discrete values, which can take the form of some number of time windows or time scales such as near future, medium future, or distant future. These are then used to directly predict the integer-valued index of future time windows of the event occurrence using (auto)regression methods~\cite{Neumann_2019_CVPR_Workshops,MINOR201777}, or to predict the ordinal values of future time scales using ordinal regression or classification~\cite{tops2013predicting}.

\noindent\textbf{2. Indirect Approaches}. These methods adopt a two-step approach, with the first step being to place the data into a series of time bins and then perform time series forecasting using techniques such as autoregressive ~\cite{bishop2006pattern} based on the historical time series $x=\{x_1,\cdots, x_T\}$ to obtain the future time series $\hat x=\{x_{T+1},\cdots, x_{\hat T}\}$. The second step is to identify events in the predicted future time series $\hat x$ using either unsupervised methods such as burstness detection~\cite{cadena2015forecasting} and change detection~\cite{kattan2015time}, or supervised techniques based on learning event characterization function. For example, existing works~\cite{8477831,1185838} first represent the predicted future time series $\hat x\in\mathbb{R}^{\hat T\times D}$ using time-delayed embedding, into $\tilde x\in\mathbb{R}^{\hat T\times D'}$ where each observation at time $t$ can be represented as $\tilde x_t=\{x_{t-(D'-1)\tau},\cdots, x_{t-2\tau},x_{t-\tau},x_t\}$ and  $t=T,T+1,\cdots \hat T$. Then an event characterization function $f_c(\tilde x_t)$ is established to map $\tilde x_t$ to the likelihood of an event, which can be fitted based on the event labels provided in the training set intuitively. Overall, the unsupervised method requires users to assume the type of patterns (e.g., burstiness and change) of future events based on prior knowledge but do not require event label data. However, in cases where the event time series pattern is difficult to assume but the label data is available, supervised learning-based methods are usually used.

\subsubsection{Continuous-time Prediction.} Discrete-time prediction methods, although usually simple to establish, also suffer from several issues. First, their time-resolution is limited to the discretization granularity; increasing this granularity significantly increases the computations al resources required, which means the resolution cannot be arbitrarily high. Moreover, this trade-off is itself a hyperparameter that is sensitive to the prediction accuracy, rendering it difficult and time-consuming to tune during training. To address these issues, a number of techniques work around it by directly predicting the continuous-valued event time~\cite{simma2012modeling}, usually by leveraging one of three techniques.

\noindent\textbf{1. Simple Regression. }The simplest methods directly formalize continuous-event-time prediction as a regression problem~\cite{bishop2006pattern}, where the output is the numerical-value future event time~\cite{wang2013towards} and/or their duration~\cite{li2017time,fu2019titan}. Common regressors such as linear regression and recurrent neural networks have been utilized  for this. Despite their apparent simplicity, this is not straightforward as simple regression typically assumes Gaussian distribution~\cite{li2017time}, which does not usually reflect the true distribution of event times. For example, the future event time needs to be left-bounded (i.e., larger than the current time), as well asbeing typically non-symmetric and usually periodic, with recurrent events having multiple peaks in the  probability density function along the time  dimension.

\noindent\textbf{2. Point Processes.} As they allow more flexibility in fitting true event time distributions, point process methods~\cite{qiao2018pairwise,weiss2013forest} are widely leveraged and have demonstrated their effectiveness for continuous time event prediction tasks. They require a conditional intensity function, defined as follows:
\begin{align}
\label{eq:intensity}
    \lambda(t|X)=\mathbb{E}[N(t,t+\mathrm{d}t)/{\mathrm{d}t}|X]=g(t|X)/(1-G(t|X))
\end{align}
where $g(t|X)$ is the conditional density function of the event occurrence probability at time $t$ given an observation $X$, and whose corresponding cumulative distribution function, $G(t|X))$, $N(t,t+\mathrm{d}t)$, denotes the count of events during the time period between $t$ and $t+\mathrm{d}t$, where $\mathrm{d}t$ is an infinitely-small time period.

Hence, by leveraging the relation between density and accumulative functions and then rearranging Equation \eqref{eq:intensity}, the following conditional density function is obtained:
\begin{align}
    g(t|X)=\lambda(t|X)\cdot\mathrm{exp}\cdot\left(-\int_{t_0}^t\lambda(u|X)\mathrm{d}u\right)
\end{align}
\indent Once the above model has been trained using a technique such as maximal likelihood~\cite{bishop2006pattern}, the time of the next event in the future is predicted as:
\begin{align}
    \hat t=\int_{t_0}^{\infty}t\cdot g(t|X)\mathrm{d}t
\end{align}

Although existing methods typically share the same workflow as that shown above, they vary in the way they define the conditional intensity function $\lambda(t|X)$. Traditional models typically utilize prescribed distributions such as the Poisson distribution~\cite{simma2012modeling}, Gamma distribution~\cite{csenki1990bayes}, Hawks~\cite{du2016recurrent}, Weibull process~\cite{damaschke2018volcanic}, and other distributions~\cite{weiss2013forest}. For example, Damaschke et al.~\cite{damaschke2018volcanic} utilized a Weibull distribution to model volcano eruption events, while Ertekin et al.~\cite{ertekin2015reactive} instead proposed the use of a non-homogeneous Poisson process to fit the conditional intensity function for power system failure events. However, in many other situations where there is no information regarding appropriate prescribed distributions, researchers must start by leveraging nonparametric approaches to learn sophisticated distributions from the data using expressive models such as neural networks. For example, Simma and Jordan~\cite{simma2012modeling} utilized of RNN to learn a highly nonlinear function of $\lambda(t|X)$.

\noindent\textbf{3. Survival Analysis}. Survival analysis~\cite{dempsey2017isurvive,vahedian2019predicting} is related to point processes in that it also defines an event intensity or hazard function, but in this case based on survival probability considerations, as follows:
\begin{align}
\label{eq:survival_analysis}
    H(t|X)=\left(\xi(t-\mathrm{d}t|X)-\xi(t|X)\right)/\xi(t|X),\ \mathrm{where }\ \xi(t|X)=p(\hat t>t)
\end{align}
where $H(t|X)$ is the so-called Hazard function denoting the hazard of event occurrence between time $(t-\mathrm{d}t)$ for a $t$ for a given observation $X$. Either $H(t|X)$ or $\xi(t|X)$ could be utilized for predicting the time of future events. For example, the event occurrence time can be estimated when $\xi(t|X)$ is lower than a specific value. Also, one can obtain $\xi(t|X)=\mathrm{exp\left(-\int_0^t H(u|X)\mathrm{d}u\right)}$ according to Equation \eqref{eq:survival_analysis}~\cite{li2007failure}. Here $H(t|X)$ can adopt any one of several prescribed models, such as the well-known Cox hazard model~\cite{li2007failure,deep2019event}. To learn the model directly from the data, some researchers have recommended enhancing it using deep neural networks~\cite{kvamme2019time}. Vahedian et al.~\cite{vahedian2019predicting} suggest learning the survival probability $\xi(t|X)$ and then applying the function $H(\cdot|X)$ to indicate an event at time $t$ if $H(t|X)$ is larger than a predefined threshold value. A classifier can also be utilized.

Instead of using the raw sequence data, the conditional intensity function can also be projected onto additional continuous-time latent state layers that eventually map to the observations~\cite{dempsey2017isurvive,vaidyanathan1999measurement}. These latent states can then be extracted using techniques such as hidden semi-Markov models~\cite{bishop2006pattern}, which ensure the elicitation of the continuous time patterns.





\subsection{Location Prediction}

\begin{figure}[htb]
  \centering
    \includegraphics[width=\textwidth]{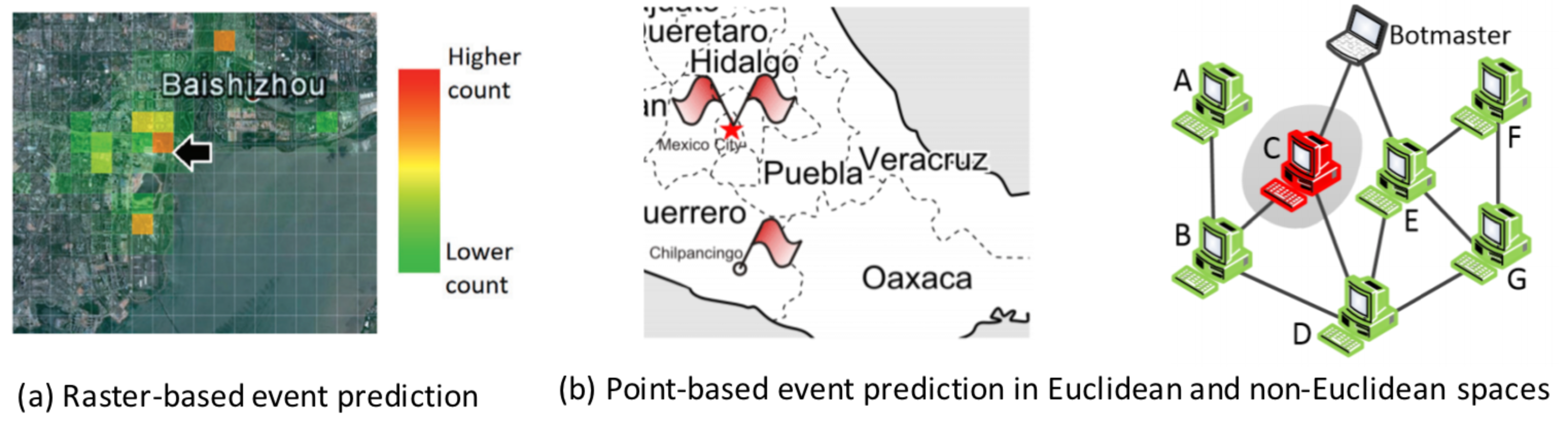}\vspace{-0.3cm}
\caption{Raster-based and Point-based Event Location Predictions.\vspace{-0.3cm}}
\label{fig:event_locations}
\end{figure}

Event location prediction focuses on predicting the location of future events. Location information can be formulated as one of two types: \textbf{1. Raster-based.} Here, a continuous space is partitioned into a grid of cells, each of which represents a spatial region, as shown in Figure \ref{fig:event_locations}(a). This type of representation is suitable for situations where the spatial size of the event is non-negligible. \textbf{2. Point-based.} In this case, each location is represented by an abstract point with infinitely-small size, as shown in Figure \ref{fig:event_locations}(b). This type of representation is most suitable for the situations where the spatial size of the event can be neglected, or the location regions of the events can only be in discrete spaces such as network nodes.

\subsubsection{Raster-based Location Prediction} There are three types of techniques used for raster-based event location prediction, namely spatial clustering, spatial embedding, and spatial convolution.

\noindent\textbf{1. Spatial clustering}. In raster-based representations, each location unit is usually a regular grid cell with the same size and shape. However, regions with similar spatial characteristics typically have irregular shapes and sizes, which could be approximated as composite representations of a number of grids~\cite{jiang2018survey}. The purpose of spatial clustering here is to group the contiguous regions who collectively exhibit significant patterns. The methods are typically agglomerative style. They typically start from the original finest-grained spatial raster units and proceed by merging the spatial neighborhood of a specific unit in each iteration. But different research works define different criteria for instantiating the merging operation. For example, Wang and Ding~\cite{wang2015hierarchical} merge neighborhoods if the unified region after merging can maintain the spatially frequent patterns. Xiong et al.~\cite{xiong2019predicting} chose an alternative approach by merging spatial neighbor locations into the current locations sequentially until the merged region possesses event data that is sufficiently statistically significant. These methods usually run in a greedy style to ensure their time complexity remains smaller than quadratic. After the spatial clustering  is completed, each spatial cluster will be input into the classifier to determine whether or not there is an event corresponding to it.

\noindent\textbf{2. Spatial interpolation}. Unlike spatial clustering-based methods, spatial interpolation based methods maintain the original fine granularity of the event location information. The estimation of event occurrence probability can be further interpolated for locations with no historical events and hence achieve spatial smoothness. This can be accomplished using commonly-used methods such as kernel density estimation~\cite{hao2019simulating,al2016area} and spatial Kriging~\cite{jiang2018survey,kleijnen2020prediction}. Kernel density estimation is a popular way to model the geo-statistics in numerous types of events such as crimes~\cite{al2016area} and terrorism~\cite{hao2019simulating}:
\begin{align}
    k(s)=1/(n\cdot \gamma)\sum\nolimits_{i=1}^n K\left((s-s_i)/\gamma\right)
\end{align}
where $k(s)$ denotes the kernel estimation for the location point $s$, $n$ is the number of historical event locations, each $s_i$ is a historical event location, $\gamma$ is a tunable bandwidth parameter, and $K(\cdot)$ is a kernel function such as Gaussian kernel~\cite{gao2019incomplete}. 

More recently, Ristea et al.~\cite{ristea2020spatial} further extended KDE-based techniques by leveraging Localized KDE and then applying spatial interpolation techniques to estimate spatial feature values for the cells in the grid. Since each cell is an area rather than a point, the center of each cell is usually leveraged as the representative of this cell. Finally, a classifier will take this as its input to predict the event occurrence for each grid~\cite{ristea2020spatial,al2016area}.

\noindent\textbf{3. Spatial convolution}. In the last few years, Convolutional Neural Networks (CNNs) have demonstrated significant success in learning and representing sophisticated spatial patterns from image and spatial data~\cite{goodfellow2016deep}. A CNN contains multiple convolutional layers that  extract the hierarchical spatial semantics of images. In each convolutional layer, a convolution operation is executed by scanning a feature map with a filter, which results in another smaller feature map with a higher level semantic. Since raster-based spatial data and images share a similar mathematical form, it is natural to leverage CNNs to process it. 

Existing methods~\cite{pirajan2019towards,mukhina2019urban,wang2019deep,bao2019spatiotemporal} in this category typically formulate a spatial map as input to predict another spatial map that denotes future event hotspots. Such a formulation is analogous to the ``image translation'' problem popular in recent years in the computer vision domain~\cite{choi2018stargan}. Specifically, researchers typically leverage an encoder-decoder architecture, where the input images (or spatial map) are processed by multiple convolutional layers into a higher-level representation, which is then decoded back into an output image with the same size, through a reverse convolutional operations process known as transposed convolution~\cite{goodfellow2016deep}.

\noindent\textbf{4. Trajectory destination prediction.} This type of method typically focuses on population-based events whose patterns can be interpreted as the collective behaviors of individuals, such as ``gathering events'' and ``dispersal events''. These methods share a unified procedure that typically consists of two steps: 1) predict future locations based on the observed trajectories of individuals, and 2) detect the occurrence of the ``future'' events based on the future spatial patterns obtained in Step 1. The specific methodologies for each step are as follows:
\begin{itemize}
    \item \textbf{Step 1: } Here, the aim is to predict each location an  individual will visit in the future, given a historical sequence of locations visited. This can be formulated as a sequence prediction problem. For example, Wang and Gerber~\cite{wang2015using} sought to predict the probability of the next time point $t+1$'s location $s_{t+1}$ based on all the preceding time points: $p(s_{t+1}|s_{\le t})=p(s_{t+1}|s_{t},s_{t-1},\cdots,s_{0})$, based on various strategies including a historical volume-based prior model, Markov models, and multi-class classification models. Vahedian et al.~\cite{vahedian2017forecasting} adopted Bayesian theory $p(s_{t+1}|s_{\le t})=p(s_{\le t}|s_{t+1})\cdot p(s_{t+1})/p(s_{\le t})$ which requires the conditional probability $p(s_{\le t}|s_{t+1})$ to be stored. However, in many situations, there is huge number of possible trajectories for each destination. For example, with a $128\times 64$ grid, one needs to store $(128\times 64)^3\approx 5.5\times 10^{11}$ options. To improve the memory efficiency, this can be limited to a consideration of just the source and current locations, leveraging a quad-tree style architecture to store the historical information. To achieve more efficient storage and speed up $p(s_{\le t}|s_{t+1})$ queries, Vahedian et al.~\cite{vahedian2017forecasting} further extended the quad-tree into a new technique called VIGO, which removes duplicate destination locations in different leaves.
    \item \textbf{Step 2: } The aim in this step is to forecast future event locations based on the future visiting patterns predicted in Step 1. The most basic strategy here is to consider each grid cell independently. For example, Wang and Gerber~\cite{wang2015using} adopted supervised learning strategies to build predictive mapping between the visiting patterns and the event occurrence. A more sophisticated approach is to consider the spatial outbreaks composited by multiple grids. Scalable algorithms have also been proposed to identify regions containing statistically significant hotspots~\cite{khezerlou2019forecasting}, such as spatial scan statistics~\cite{kulldorff1997spatial}. Khezerlou et al.~\cite{khezerlou2019forecasting} proposed a greedy-based heuristic tailored for the grid-based data formulation, which extends the original ``seed'' grid containing statistically-large future event densities to four directions until the extended region is no longer a statistically-significant outbreak.
\end{itemize}

\subsubsection{Point-based Prediction}. Unlike the raster-based formulation, which covers the prediction of a contiguous spatial region, point-based prediction focuses specifically on locations of interest, which can be distributed sparsely in a Euclidean (e.g., spatial region) or non-Euclidean space (e.g., graph topology). These methods can be categorized into supervised and unsupervised approaches.

\noindent\emph{1. Supervised approaches.}
In supervised methods, each location will be classified as either ``positive'' or ``negative'' with regard to a future event occurrence. The simplest setting is based on the independent and identically distributed (i.i.d.) assumption among the locations, where each location is predicted by a classifier independently using their respective input features. However, given that different locations usually have strong spatial heterogeneity and dependency, further research has been proposed to tackle them based on different locations' predictors and outputs, resulting in two research directions: 1) Spatial multi-task learning, and 2) Spatial auto-regressive methods.
\begin{itemize}
\item\textbf{Spatial multi-task learning.} Multi-task learning is a popular learning strategy that can jointly learn the models for different tasks such that the learned model can not only share their knowledge but also preserve some exclusive characteristics of the individual tasks~\cite{zhao2015multi}. This notion coincides very well with spatial event prediction tasks, where combining the outputs of models from different locations needs to consider both their spatial dependency and heterogeneity. Zhao et al.~\cite{zhao2015multi} proposed a spatial multi-task learning framework as follows:
\begin{align}
    \min_{W}\sum\nolimits_i^m\mathcal{L}(Y_i, f(W_i,X_i))+\alpha\cdot\mathcal{R}(\{W_i\}_i^k,M),\ \  \ s.t.,\ \mathcal{C}(\{W_i\}_i^m,D)\in\mathbb{C}
\end{align}
where $m$ is the total number of locations (i.e., tasks), $W_i$ and $Y_i$ are the model parameters and true labels (event occurrence for all time points), respectively, of task $i$. $\mathcal{L}(\cdot)$ is the empirical loss, $f(W_i,X_i)$ is the predictor for task $i$, and $\mathcal{R}(\cdot)$ is the spatial regularization term based on the spatial dependency information $M\in\mathbb{R}^{m\times m}$, where $M_{i,j}$ records the spatial dependency between location $i$ and $j$. $\mathcal{C}(\cdot)$ represents the spatial constraints imposed over the corresponding models to enforce them to remain within the valid space $\mathbb{C}$. Over recent years, there have been multiple studies proposing different strategies for $\mathcal{R}(\cdot)$ and $\mathcal{C}(\cdot)$. For example, Zhao et al.~\cite{zhao2017feature} assumed that all the locations would be evenly correlated and enforced their similar sparsity patterns for feature selection, while Gao et al.~\cite{gao2019incomplete} further extended this to differentiate the strength of the correlation between different locations' tasks according to the spatial distance between them. This research has been further extended this approach to tree-structured multitask learning to handle the hierarchical relationship among locations at different administrative levels (e.g., cities, states, and countries)~\cite{zhao2017spatial} in a model that also considers the logical constraints over the predictions from different locations who have hierachical relationships. Instead of evenly similar, Zhao, et al.~\cite{zhao2019spatial} further estimated spatial dependency $D$ utilizing inverse distance using Gaussian kernels, while Ning et al.~\cite{ning2018staple} proposed estimating the spatial dependency $D$ based on the event co-occurrence frequency between each pair of locations.

\item\textbf{Spatial auto-regressive methods}. Spatial auto-regressive models have been extensively explored in domains such as geography and econometrics, where they are applied to perform predictions where the i.i.d. assumption is violated due to the strong dependencies among neighbor locations. Its generic framework is as follows:
\begin{align}
    \hat Y_{t+1}=\rho M \hat Y_{t+1}+X_t\cdot W+\varepsilon
\end{align}
where $X_t\in\mathbb{R}^{m\times D}$ and $\hat Y_{t+1}\in\mathbb{R}^{m\times m}$ are the observations at time $t$ and event predictions at time $t+1$ over all the $m$ locations, and $M\in\mathbb{R}^{m\times m}$ is the spatial dependency matrix with zero-valued diagonals. This means the prediction of each location $\hat Y_{t+1,i}\in \hat Y_{t+1}$ is jointly determined by its input $X_{t,i}$ and neighbors $\{j|M_{i,j}\ne 0\}$  and $\rho$ is a positive value to balance these two factors. Since event occurrence requires discrete predictions, simple threshold-based strategies can be used to discretize $\hat Y_i$ into $\hat Y_i'=\{0,1\}$~\cite{calabrese2016estimating}. Moreover, due to the complexity of event prediction tasks and the large number of locations, sometimes it is difficult to define the whole $M$ manually. Zhao et al.~\cite{zhao2019spatial} proposed jointly learning the prediction model and spatial dependency from the data using graphical LASSO techniques. Yi et al.~\cite{yi2018integrated} took a different approach,  leveraging conditional random fields to instantiate the spatial autoregression, where the spatial dependency is measured by Gaussian kernel-based metrics. Yi et al.~\cite{yi2019neural} then went on to propose leveraging the neural network model to learn the locations' dependency.
\end{itemize}
\noindent\emph{2. Unsupervised approaches}. Without supervision from labels, unsupervised-based methods must first identify potential precursors and determinant features in different locations. They can then detect anomalies that are characterized by specific feature selection and location combinatorial patterns (e.g., spatial outbreaks and connected subgraphs) as the future event indicators~\cite{chen2014non}. The generic formulation is as follows:
\begin{align}
\label{eq:scan}
    (F,R)=\arg\max\nolimits_{F,R}\  q(F,R)\ \ \  s.t.,\ \mathrm{supp}(F_i)\in\mathbb{M}(\mathbb{G},\beta),\ \mathrm{supp}(R_i)\in\mathbb{C}
\end{align}
where $q(\cdot)$ denotes scan statistics which score the significance of each candidate pattern, represented by both a candidate location combinatorial pattern $R$ and feature selection pattern $F$. Specifically, $F\in\{0,1\}^{D'\times n}$ denotes the feature selection results (where ``1'' means selected; ``0'', otherwise) and $R\in\{0,1\}^{m\times n}$ denotes the $m$ involved locations for the $n$ events. $\mathbb{M}(\mathbb{G},\beta)$ and $\mathbb{C}$ are the set of all the feasible solutions of $F$ and $R$, respectively. $q(\cdot)$ can be instantiated by scan statistics such as Kulldorff's scan statistics~\cite{kulldorff1997spatial} and the Berk-Jones statistic~\cite{chen2014non}, which can be applied to detect and forecast events such as epidemic outbreaks and civil unrest events~\cite{ramakrishnan2014beating}. Depending on whether the embedding space is an Euclidean region (e.g., a geographical region) or a non-Euclidean region (e.g., a network topology), the pattern constraint $\mathcal{C}$ can be either constrained to predefined geometric shapes such as a circle, rectangle, or an irregular shape or subgraphs such as connected, cliques, and k-cliques. The problem in Equation \eqref{eq:scan} is nonconvex and sometimes even discrete, and hence difficult to solve. A generic way is to optimize $F$ using sparse feature selection; there is a useful survey provided in~\cite{li2017feature} and $R$  can be defined  using the two-step graph-structured matching method detailed in\cite{chen2017generic}. More recently, new techniques have been developed that are capable of jointly learning both feature and location selection~\cite{chen2017generic,shao2017efficient}.

\subsection{Semantic Prediction} Event semantics prediction addresses the problem of forecasting topics, descriptions, or other meta-attributes in addition to future events' times and locations. Unlike time and location prediction, the data in event semantics prediction usually involves symbols and natural languages in addition to numerical quantities, which means different types of techniques may be utilized. The data are categorized into three types based on how the  historical data are organized and utilized to infer future events. The first of these categories covers rule-based methods, where future event precursors are extracted by mining association or logical patterns in historical data. The second type is sequence-based, considering event occurrence to be a consequence of temporal event chains. The third type further generalizes event chains into event graphs, where additional cross-chain contexts need to be modeled. These are discussed in turn below.

\subsubsection{Association-based Prediction}.
Association rule-based methods are amongst the most classic approaches in data mining domain for event prediction, typically consisting of two steps: 1) learn the associations between precursors and target events, and then 2) utilize the learned associations to predict future events. For the first step, for example, an association could be $x=\{$``election'', ``fraud''$\}\rightarrow\ y=$``protest event'', which indicates that serious fraud occurring in an election process could lead to future protest events. To discover all the significant associations from the ocean of candidate rules efficiently, frequent set mining~\cite{han2011data} can be leveraged. Each discovered rule needs to come with both sufficient \emph{support} and \emph{confidence}. Here, \emph{support} is defined as the number of cases where both ``$x$'' and ``$y$'' co-occur, while \emph{confidence} means the ratio indicating that ``$y$'' occurs once ``$x$'' happens. To better estimate these discrimination rules, further temporal constraints can be added that require the occurrence time of ``$x$'' and ``$y$'' to be sufficiently close to be considered ``co-occurrences''. Once the frequent set rules have been discovered, pruning strategies may be applied to retain the most accurate and specific ones, with various strategies for generating final predictions~\cite{han2011data}. Specifically, given each new observation $x'$, one of the simplest strategies is to output the events that are triggered by any of the association rules starting from event $x'$~\cite{vilalta2002predicting}. Other strategies first rank the predicted results based on their confidence and then predict just the top $r$ events~\cite{zhou2015pattern}. More sophisticated and rigorous strategies tend to build a decision list where each element in the list is an association rule mapping, so once a generative model has been built for the decision process, the maximal likelihood can be leveraged to optimize the order of the decision list~\cite{letham2015interpretable}.

\subsubsection{Causality-based prediction}
This type of research leverages the causality inferred among the historical events to achieve future event predictions. The  data here typically shares a generic framework consisting of the following procedures: 1) event representation, 2) event graph construction, and 3) future event inference.

\noindent\textbf{Step 1: Event semantic representation. }This approach typically begins by extracting the events from the target texts using natural language processing techniques such as sanitization, tokenization, POS tag analysis, and name entity recognition. Several types of objects can be extracted to represent the events: i) Noun Phrase-based~\cite{chan2005extracting,hashimoto2014toward,khoo2000extracting}, where the noun-phrase corresponds to each event (for example, ``2008 Sichuan Earthquake''); ii) Verbs and Nouns~\cite{kim1993supervenience,radinsky2012learning}, where an event is represented as a set of noun-verb pairs extracted from news headlines (for example, ``<capture, people>'', ``<escape, prison>'', or ``<send, prison>''); and iii) Tuple-based~\cite{zhao2017constructing}, where each event is represented by a tuple consisting of objects (such as actors, instruments, or receptors), a relationship (or property), and time. An RDF-based format has also been leveraged in some works~\cite{dami2018news}.

\noindent\textbf{Step 2: Event causality inference. } The goal here is to infer the cause-effect pairs among historical events. Due to its combinatorial nature, narrowing down the number of candidate pairs is crucial. Existing works usually begin by clustering the events into event chains, each of which consist of a sequence of time-ordered events under the relevant semantics, typically the same topics, actors, and/or objects~\cite{acharya2017causal}. The causal relations among the event pairs can then be inferred in various ways. The simplest approach is just to consider the likelihood that $y$ occurs after $x$ has occurred throughout the training data. Other methods utilize NLP techniques to identify causal mentions such as causal connectives, prepositions, and verbs~\cite{radinsky2012learning}. Some formulate causal-effect relationship identification as a classification task where the inputs are the cause and effect candidate events, often incorporating contextual information including related background knowledge from web texts. Here, the classifier is built on a multi-column CNN that outputs either ``1'' or ``0'' to indicate whether the candidate has an effect or not~\cite{kruengkrai2017improving}. In many situations, the cause-effect rules learned directly using the above methods can be too specific and sparse, with low generalizability, so a typical next step is to generalize the learned rules. For example, ``Earthquake hits China'' $\rightarrow$ ``Red Cross help sent to Beijing'' is a specific rule that can be generalized to ``Earthquake hits [A country]'' $\rightarrow$ ``Red Cross help sent to [The capital of this country]''. To achieve this, some external ontology or a knowledge base is typically needed in order to establish the underlying relationships among items or provide necessary information on their properties, such as Wikipedia (\url{https://www.wikipedia.org/}), YAGO~\cite{suchanek2007yago}, WordNet~\cite{fellbaum2012wordnet}, or ConceptNet~\cite{liu2004conceptnet}. Based on these resources, the similarity between two cause-effect pairs $(c_i,\varepsilon_i)$ and $(c_j,\varepsilon_j)$ can be computed by jointly considering the respective similarity of the putative cause and effect: $\sigma((c_i,\varepsilon_i),(c_j,\varepsilon_j))=(\sigma(c_i,c_j)+\sigma(\varepsilon_i,\varepsilon_j))/2$. An appropriate algorithm can then be utilized to apply hierarchical agglomerative clustering to group them and hence generate a data structure that can efficiently manage the task of storing and querying them to identify any cause-effect pairs. For example,~\cite{shrestha2017predicting,radinsky2012learning,radinsky2012learning_J} leverage an abstraction tree, where each leaf is an original specific cause-effect pair and each intermediate node is the centroid of a cluster. Instead of using hierarchical clustering,~\cite{zhao2017constructing} directly uses the word ontology to simultaneously generalize cause and effect (e.g., the noun ``violet'' is generalized to ``purple'', the verb ``kill'' is generalized to ``murder-42.1\footnote{the form of verb class in VerbNet~\cite{zhao2017constructing}.}'') and then leverage a hierarchical causal network to organize the generalized rules.


\noindent\textbf{Step 3: Future event Inference. } Given an arbitrary query event, two steps are needed to infer the future events caused by it based on the causality of events learned above. First, we need to retrieve similar events that match the query event from historical event pool. This requires the similarity between the query event and all the historical events to be calculated. To achieve this, Lei et al.~\cite{lei2019event} utilized context information, including event time, location, and other environmental and descriptive information. For methods requiring event generalization, the first step is to traverse the abstraction tree starting from the root that corresponds to the most general event rule. The search frontier then moves across the tree if the child node is more similar, culminating in the nodes which are the least general but still similar to the new event being retrieved~\cite{radinsky2012learning}. Similarly,~\cite{cho2008tree} proposed another tree structure referred to as a ``circular binary search tree'' to manage the event occurrence pattern. We can now apply the learned predicate rules starting from the retrieved event to obtain the prediction results. Since each cause event can lead to multiple events, a convenient way to determine the final prediction is to calculate the support~\cite{radinsky2012learning}, or conditional probability~\cite{yang2019using} of the rules. Radinsky et al.~\cite{radinsky2012learning} took a different approach, instead ranking the potential future events by their similarity defined by the length of their minimal generalization path. For example, the minimal
generalization path for ``London'' and ``Paris'' is ``London''$\xrightarrow{\text{capital-of}}$``Great Brain''$\xrightarrow{\text{in-continent}}$``Europe''$\xleftarrow{\text{in-continent}}$``France''$\xleftarrow{\text{capital-of}}$``Paris''. Alternatively, Zhao et al.~\cite{zhao2017constructing} proposed embedding the event causality network into a continuous vector space and then applying an energy function designed to rank potential events, where true cause-effect pairs are assumed to have low energies.



\subsubsection{Semantic Sequence} These methods share a very straightforward problem formulation. Given a temporal sequence for a historical event chain, the goal is to predict the semantics of the next event using sequence prediction~\cite{bishop2006pattern}. The existing methods can be classified into four major categories: 1) classical sequence prediction; 2) recurrent neural networks; 3) Markov chains; and 4) time series predictions.

\noindent\textbf{Sequence classification-based methods}. These methods formulate event semantic prediction as a multi-class classification problem, where a finite number of candidate events are ranked and the top-ranked event is treated as the future event semantic. The objective is $\hat C=\arg\max_{C_i} u(s_{T+1}=C_i|s_1,\cdots,s_T)$, where $s_{T+1}$ denotes the event semantic in time slot $T+1$ and $\hat C$ is the optimal semantic among all the semantic candidates $C_i$ ($i=1,\cdots$). Multi-class classification problems can be split into events with different topics/semantic meaning. Three types of sequence classification methods have been utilized for this purpose, namely feature-based methods, prototype-based methods, and model-based methods such as Markov models. 
\begin{itemize}
    \item \textbf{Feature-based.} One of the simplest methods is to ignore the temporal relationships among the events in the chain, by either aggregating the inputs or the outputs. Tama and Comuzzi~\cite{tama2019empirical} formulated historical event sequences with multiple attributes for event prediction, testing multiple conventional classifiers. Another type of approach based on this notion utilizes compositional based-methods~\cite{granroth2016happens} that typically leverage the assumption of independency among the historical input events to simplify the original problem $u(s_{T+1}|s_1,s_2,\cdots,s_T)=u(s_{T+1}|s_{\le T})$ into $v(u(s_{T+1}|s_1),u(s_{T+1}|s_2),\cdots,u(s_{T+1}|s_T))$ where $v(\cdot)$ is simply an aggregation function that represents a summation operation over all the components. Each component function $u(s_{T+1}|s_i)$ can then be calculated by estimating how likely it is that event semantic $s_{T+1}$ and $s_i$ ($i\le T$) co-occur in the same event chain. Granroth-Wilding and Clark~\cite{granroth2016happens} investigated various models ranging from straightforward similarity scoring functions through bigram models and word embedding combined with similarity scoring functions to newly developed composition neural networks that jointly learn the representation of $s_{T+1}$ and $s_i$ and then calculate their coherence. Some other researchers have gone further to consider the dependency among the historical events. For example, Letham et al.~\cite{letham2013sequential} proposed to optimizing the correct ordering among the candidate events, based on the following equation:
\begin{align}\small\nonumber
    \sum\nolimits_{i\in\mathcal I,\ j\in\mathcal J} \mathbbm{1}_{[u(s_{T+1}=C_i|s_{\le T})>u(s_{T+1}=C_j|s_{\le T})]}\implies \sum\nolimits_{i\in\mathcal I,\ j\in\mathcal J} e^{u(s_{T+1}=C_i|s_{\le T})-u(s_{T+1}=C_j|s_{\le T})}+\rho\|W\|_2^2
\end{align}\normalsize
where the semantic candidate in the set $\mathcal I$ should be ranked strictly to be lower than those in $\mathcal J$, with the goal being to penalize the ``incorrect ordering''. Here, $\mathbbm{1}_{[\cdot]}$ is an indicator function which is discrete such that $\mathbbm{1}_{[b\ge a]}\le e^{b-a}$ and can thus be utilized as the upper-bound for minimization, as can be seen in the right-hand-side of the above equation. $W$ is the set of parameters of the function $u(\cdot)$. This can now be relaxed to an exponential-based approximation for effective optimization using gradient-based algorithms~\cite{goodfellow2016deep}. Other methods focus on first transferring the sequential data into sequence embeddings that can encode the latent sequential context. For example, Fronza et al.~\cite{fronza2013failure} apply random indexing to represent the words in  terms of their its vector representations by embedding the information from neighboring words into each word before utilizing conventional classifiers such as Support Vector Machines (SVM) to identify the future events.
\item\textbf{Model-based.} Markov-based models have also been leveraged to characterize temporal patterns~\cite{yang2014finding}. 
These typically use $E_i$ to denote each event under a specific type and $\mathcal{E}$ denotes the set of event types. The goal here is to predict the event type of the next event to occur in the future. In~\cite{alevizos2017event}, the event types are modeled using the Markov model so given the current event type, the next event type can be inferred simply by looking up the state with the highest probability in the transition matrix. A tool called Wayeb~\cite{alevizos2018wayeb} has been developed based on this method. Laxman et al.~\cite{laxman2008stream} developed a more complicated model, based on a mixture of Hidden Markov models and introducing new assumptions and the concept of episodes composed of a subsequence of event types. They assumed different event episodes should have different transition patterns so started by discovering the frequent episodes for events, each of which they modeled by a specific  hidden Markov model over various event types. This made it possible to establish the generative process for each future event type $s$ based on the mixture of the above episode Markov models. When predicting, the likelihood of a current observed event sequence over each possible generative process, $p(X|\Lambda_Y)$ is evaluated, after which a future event type can be considered as either being larger than some threshold (as in~\cite{laxman2008stream}) or the largest among all the different $Y$ values (in~\cite{zhao2015spatiotemporal,zhao2016online}). 
\item\textbf{Prototype-based.} Adhikari et al.~\cite{adhikari2019epideep} took a different approach, utilizing a prototype-based strategy that first clusters the event sequences into different clusters in terms of their temporal patterns. When a new event sequence is observed, its closest cluster's centroid will then be leveraged as a ``reference event sequence'' whose sub-sequential events will be referred to when predicting future events for this new event sequence.
\end{itemize}

\noindent\textbf{Recurrent neural network (RNN)-based methods}. Approaches in this category can be classified into two types: 1. Attribute-based models; and 2. Descriptive-based models. The attribute-based models, ingest feature representation of events as input, while the descriptive-based models typically ingest unstructured information such as texts to directly predict future events.
\begin{itemize}
\item\textbf{Attributed-based Methods. }Here, each event $y=(t,l,s)$ at time $t$ is recast and represented as $e_t=(e_{t,1},e_{t,2},\cdots,e_{t,k})$, where $e_{t,i}$ is the $i$-th feature of the event at time $t$. The feature here can include location and other information such as event topic and semantics. Each sequence $e=(e_1,\cdots,e_t)$ is then input into the standard RNN architecture for predicting next event $e_{t+1}$ in the sequence at time point $t+1$~\cite{lin2019mm}. Various types of RNN components and architecture have been utilized  for this purpose~\cite{casagrande2018sensor,casagrande2019prediction}, but a vanilla RNN~\cite{goodfellow2016deep,duan2019clinical} for sequence-based event prediction can be written in the following form:
\begin{align}\nonumber
   \indent  h_i=\mathrm{tahn}(a_{t}),\ a_{i}=b+W\cdot h_{i-1}+U\cdot e_i,\ o_i=c+V\cdot h_i,\ \psi(i+1)=\mathrm{softmax}(o_i),\ i\le t
\end{align}
where $h_i$, $o_i$, and $a_i$ are the latent state, output, and activation for the $i$-th event, respectively, and $W$, $U$, and $V$ are the model parameters for fitting the corresponding mappings. The prediction $e_{t+1}:=\psi(t+1)$ can then be calculated in a feedforward way from the first event and the model training can be done by back-propagating the error from the layer of $\psi(t)$. Existing work typically utilizes the variants of vanilla RNN to handle the gradient vanishing problem, especially when the event chain is not short. The most commonly used methods for event prediction are LSTM and GRU~~\cite{goodfellow2016deep}. For example, the architecture and equation for LSTM are as follows:
\begin{align}
    &a_i=\sigma(W_j\cdot[h_{i-1},e_i]+b_j),\ \tilde C_i=\mathrm{tanh}(W_C\cdot[h_{i-1},h_i]+b_C),\ C_i=\zeta_i C_{i-1}+a_i\tilde C_i,\\\nonumber
    &\ \ \ \ \zeta_i=\sigma(W_\zeta\cdot[h_{i-1},e_i]+b_\zeta),\ o_i=\sigma(W_o[h_{i-1}],e_i)+b_o),\ h_i=o_i*\mathrm{tahn}(C_i)
\end{align}
where the additional components $C_{i-1}$ and $\zeta_i$ are introduced to keep tracking the previous ``history'' and gating the information for forgetting in order to handle longer sequences. For example, some researchers opt to leverage a simple type LSTM architecture to extend the RNN-based sequential event prediction~\cite{casagrande2018sensor,hu2017happens}, while others leverage variants of LSTM, such as bi-directional LSTM instead~\cite{nguyen2017sequence,kiyomaru2019diversity}and yet others prefer to leverage gated-recurrent units (GRU)~\cite{duan2019clinical}.

Moving beyond considering just the chain relationships among events, Li et al.~\cite{li2018constructing} generalized this into graph-structured relationships to better incorporate the event contextual information via the Narrative Event Evolutionary Graph (NEEG). An NEEG is a knowledge graph where each node is an event and each edge denotes the association between a pair of events, enabling the NEEG to be represented by  a weighted adjacency matrix $A$. The basic architecture can be denoted by the following, as detailed in the paper~\cite{li2018constructing}:
\begin{align}
    &a_i=A^{\mathrm{T}}h_{i-1}+b,\ z_i=\sigma(W_z a_i+U_z h_{i-1}),\ r_i=\sigma(W_r a_i+U_r h_{i-1}),\\\nonumber
    &\ \ \ c_i=\mathrm{tanh}(W a_i+U(r_i  h_{i-1})),\ h_i=(1-z_i)h_{i-1}+z_i c_i
\end{align}
Here, the current activation $a_i$ is not only dependent on the previous time point but also influenced by its neighbor nodes in NEEG.

\item\textbf{Descriptive-based Methods. }Attribute-based methods require extra effort during pre-processing in order to convert the unstructured raw data into feature vectors, a process which is not only computationally labor intensive but also not always feasible. Therefore, multiple architectures have been proposed to directly process the raw (textual) event descriptions to enable them to be used to predict future event semantics or descriptions. These models share a similar generic framework~\cite{hu2017happens,xue2018miml,lv2019sam,hu2019integrating,yu2019dram,su2020hierarchical}, which begins by encoding each sequence of words into event representations, utilizing an RNN architecture, as shown in Figure~\ref{figure:descriptive_RNN_framework}. The sequence of events must then be characterized by another higher-level RNN to predict future events. Under this framework, some works begin by decoding the predicted future candidate events into event embedding, after which they are compared with each other and the one with the largest confidence score is selected as the predicted event. These methods are usually constrained by the known list of event types, but sometimes we are interested in open set predictions where the predicted event type can be a new appearance of a type that has not previously been seen in the training set. To achieve this, other methods focus on directly generating future events' descriptions that characterize event semantics that may or may not have appeared before by designing an additional sequence decoder that decodes the latent representation of future events into word sequences. More recent research has enhanced the utility and interpretability of the relationship between words and relevant events, and all the previous events for the relevant future event, by adding a hierarchical attention mechanisms. For example, Yu et al.~\cite{yu2019dram} and Su and Jiang~\cite{su2020hierarchical} both proposed word-level attention and event-level attention, while Hu~\cite{hu2019integrating} leveraged word-level attention in the event encoder as well as in the event decoder. 
\begin{figure}[htb]
  \centering
    \includegraphics[width=0.8\textwidth]{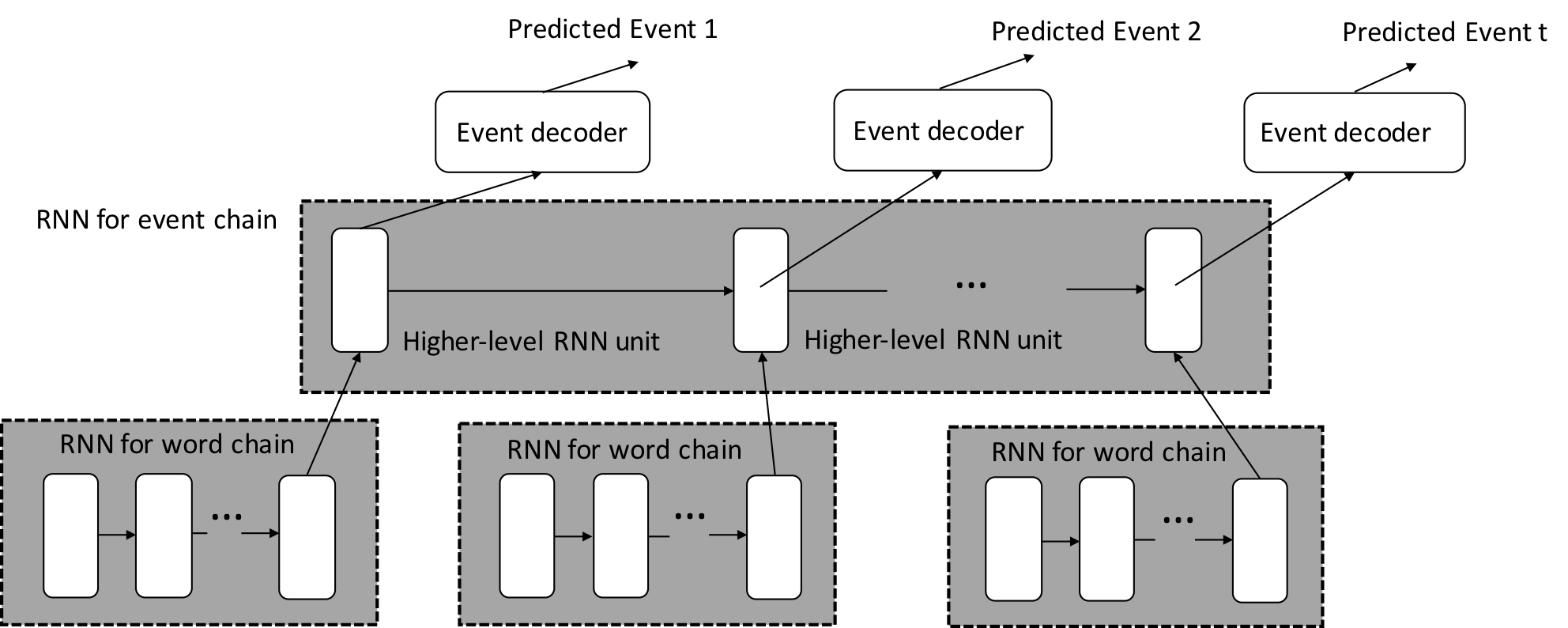}\vspace{-0.3cm}
\caption{Generic framework for hierarchical RNN-based event forecasting.\vspace{-0.3cm}}
\label{figure:descriptive_RNN_framework}
\end{figure}
\end{itemize}







\subsection{Multi-faceted Prediction} This section discusses the research into ways to jointly predict the time, location, and semantics of future events. Existing work in this area can be categorized into three types: 1) joint time and semantics prediction; 2) joint time and location prediction; and 3) joint time, location, and semantic prediction.
\subsubsection{Time and Semantics}. For joint time and semantic prediction, there are three popular types of methods, discussed in turn below


\noindent\textbf{Temporal association rule}. A temporal association rule can be developed from the vanilla association rule $LHS\rightarrow RHS$ by embedding additional temporal information into either $LHS$, $RHS$  or both, thus redefining the meaning of co-occurrence and association with temporal constraints. For example, Vilalta and Ma~\cite{vilalta2002predicting} defined $LHS$ as a tuple $(E_L,\tau)$, where $\tau$ is the time window before the target in $RHS$ predefined by the user. Only the events occurring within a time window before the event in $RHS$ will satisfy the $LHS$. Similar techniques have also been leveraged by other researchers~\cite{srinivasa2008high,cho2008tree}. However, $\tau$ is difficult to define beforehand and it is preferable to be flexible to suit different target events. To handle this challenge, Yang et al.~\cite{yang2002web} proposed a way to automatically identify information on a continuous time interval from the data. Here, each transaction is composed of not only items but also continuous time duration information. $LHS$ is a set of items (e.g., previous events) while $RHS$ is a tuple $(E_R,[t_1,t_2])$ consisting of a future event semantic representation and its time interval of occurrence. To automatically learn the time interval in $RHS$,~\cite{yang2002web} proposed the use of two different methods . The first is called the \emph{confidence-interval-based} method, which leverages a statistical distribution (e.g., Gaussian and student-t~\cite{bishop2006pattern}) to fit all the observed occurrence times of events in $RHS$, and then treats the statistical confidence interval as the time interval. The second method is known as \emph{minimal temporal region selection}, which aims to find the temporal region with the smallest interval and covers all historical occurrences of the event in $RHS$.


\noindent\textbf{Time expression extraction}. In contrast to the above statistical-based methods, another way to achieve event time and semantics joint prediction comes from the pattern recognition domain, aiming to directly discover time expressions that mention the (planned) future events. As this type of technique can simultaneously identify time, semantics, and other information such as locations, it is widely used and will be discussed in more details later as part of the discussion of ``Planned future event detection methods'' in Section \ref{sec:time_location_semantics}.

\begin{figure}[htb]
  \centering
    \includegraphics[width=0.8\textwidth]{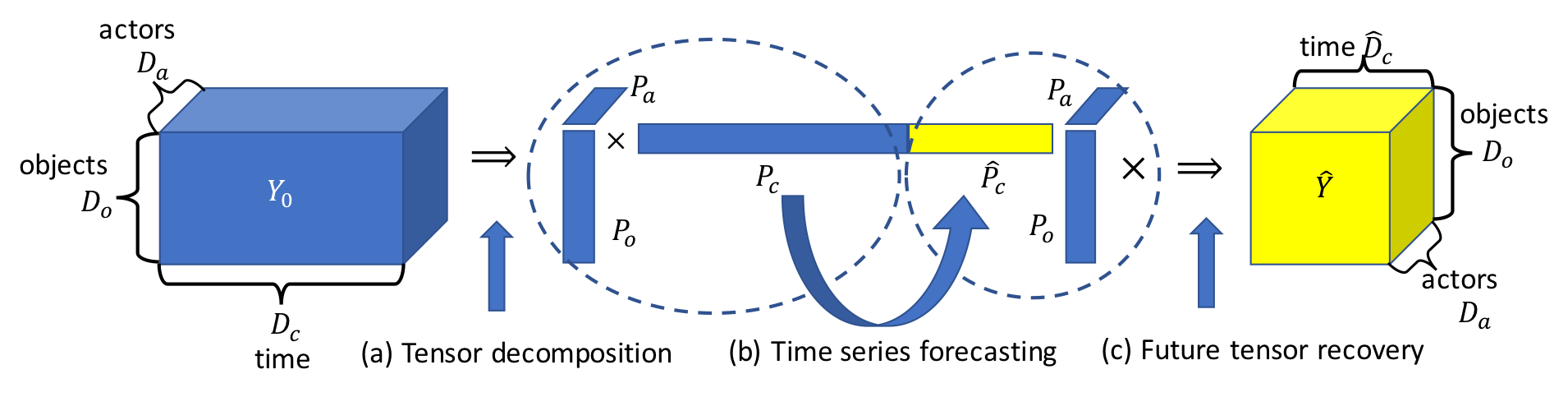}\vspace{-0.3cm}
\caption{Tensor Decomposition and Forecasting for Complex Time-stamped events.\vspace{-0.3cm}}
\label{figure:tensor_decomposition}
\end{figure}

\noindent\textbf{Time series forecasting-based methods. }The methods based on time series forecasting can be separated into direct methods and indirect methods. \emph{Direct methods} typically formulate the event semantic prediction problem as a multi-variate time series forecasting problem, where each variable corresponds to an event type $C_i\ (i=1,\cdots)$ and hence the predicted event type at future time $\hat t$ is calculated as $\hat s_{\hat t}=\arg\max_{C_i} f(s_{\hat t}=C_i|X)$. For example, in~\cite{li2017multi}, a longitudinal support vector regressor is utilized to predict multi-attribute events, where $n$ support vector regressors, each of which corresponds to an attribute, is built to achieve the goal of predicting the next time point's attribute value. Weiss and Page~\cite{weiss2013forest} took a different approach, leveraging multiple point process models to predict multiple event types. To further estimate the confidence of their predictions, Bilo{\v{s}} et al.~\cite{bilovs2019uncertainty} first leveraged RNN to learn the historical event representation and then input the result into a Gaussian process model to predict future event types. To better capture the joint dynamics across the multiple variables in the time series, Brandt et al.~\cite{brandt2011real} extended this to Bayesian vector autoregression. Utilizing \emph{indirect-style} methods, they focused on learning a mapping from the observed event semantics down to low-dimensional latent-topic space using tensor decomposition-based techniques. Similarly, Matsubara et al.~\cite{matsubara2012fast} proposed a 3-way topic analysis of the original observed event tensor $Y_0\in\mathbb{R}^{D_o\times D_a\times D_c}$ consisting of three factors, namely actors, objects, and time. They then went on to decompose this tensor into latent variables via three corresponding low-rank matrices $P_{o}\in \mathbb{R}^{D_k\times D_o}$, $P_{a}\in \mathbb{R}^{D_k\times D_a}$, and $P_{c}\in \mathbb{R}^{D_k\times D_c}$ respectively, as shown in Figure~\ref{figure:tensor_decomposition}. Here $D_k$ is the number of latent topics. For the prediction, the time matrices $P_{c}$ are predicted into the future $\hat P_{c}$ via multi-variate time series forecasting, after which a future event tensor are estimated by recovering a ``future event tensor'' $\hat Y$ by the multiplication among the predicted time matrix $\hat P_{c}$ as well as the known actor matrix $P_{a}$ and object matrix $P_{o}$.

\subsubsection{Time and Location Prediction} This category of methods focuses on jointly predicting the location and time of future events. These methods can be classified into two subtypes: 1) raster-based: which focus on predictions for individual time slots and location regions, and 2) point-based: which predicts continuous time and location points.

\noindent\textbf{Raster-based}. These methods usually formulate data into temporal sequences consisting of spatial snapshots. Over the last few years, various techniques have been proposed to characterize the spatial and temporal information for event prediction.

The simplest way to consider spatial information is to directly treat location information as one of the input features, and then feed it into predictive models, such as linear regression~\cite{zhao2017modeling}, LSTM~\cite{ren2018deep} and Gaussian processes~\cite{kupilik2018spatio}. During model training, Zhao and Tang~\cite{zhao2017modeling} leveraged the spatiotemporal dependency to regularize their model parameters.  
\begin{figure}[htb]
  \centering
    \includegraphics[width=0.9\textwidth]{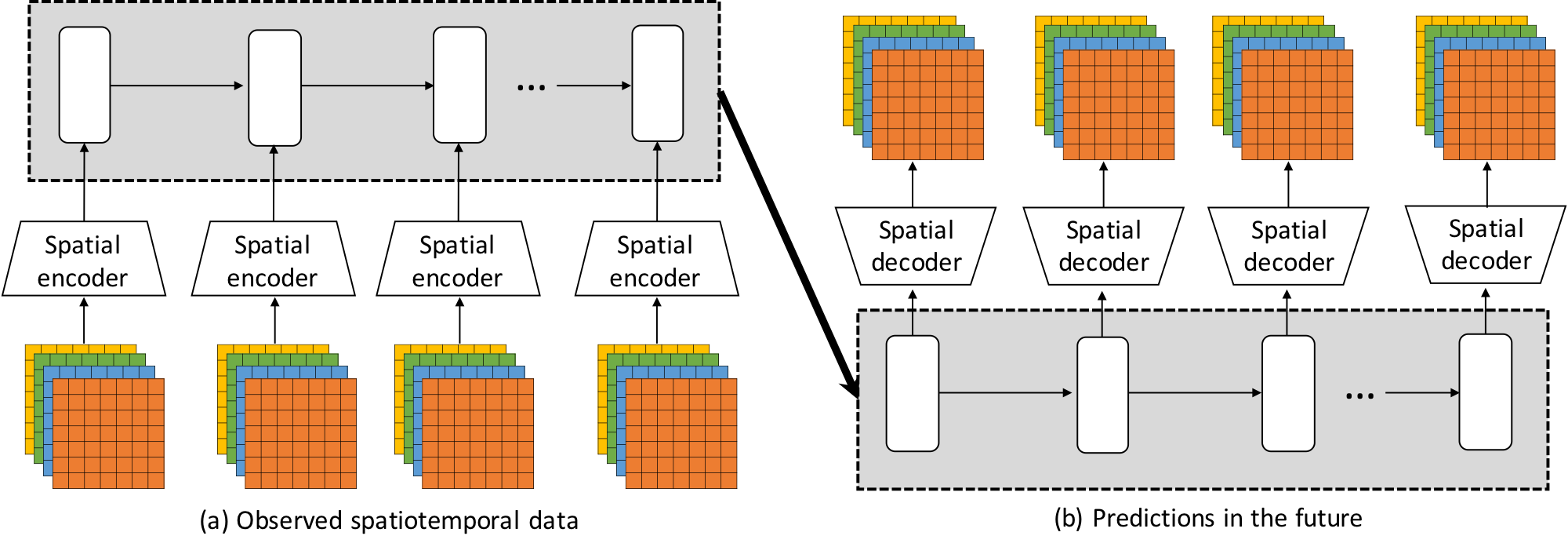}\vspace{-0.3cm}
\caption{Generic framework for spatiotemporal event prediction using CNN+RNN-based deep learning framework.\vspace{-0.3cm}}
\label{figure:cnn_rnn}
\end{figure}
Most of the methods in this domain aim to jointly consider the spatial and temporal dependency for predictions~\cite{di2019traffic}. At present, the most popular framework is the CNN+RNN architecture, which implements sequence-to-sequence learning problems such as the one illustrated in Figure \ref{figure:cnn_rnn}. Here, the multi-attributed spatial information for each time point can be organized as a series of multi-channel images, which can be encoded using convolution-based operations. For example, Huang et al.~\cite{huang2019mist} proposed the addition of convolutional layers to process the input into vector representations. Other researchers have leveraged variational autoencoders~\cite{wang2020csan} and CNN autoencoders~\cite{jiang2019deepurbanevent} to learn the low-dimensional embedding of the raw spatial input data. This allows the learned representation of the input to be input into the temporal sequence learning architecture. Different recurring units have been investigated, including  RNN,  LSTM, convLSTM, and stacked-convLSTM~\cite{goodfellow2016deep}. The resulting representation of the input sequence is then sent to the output sequence as input. Here, another recurrent architecture is established. The output of the unit for each time point will be input into a spatial decoder component which can be implemented using transposed convolutional layers~\cite{yuan2018hetero}, transposed convLSTM~\cite{jiang2019deepurbanevent}, or a spatial decoder in a variational autoencoder~\cite{wang2020csan}. A conditional random field is another popular technique often used to model the spatial dependency~\cite{jiang2018survey}.

\noindent\textbf{Point-based}. The spatiotemporal point process is an important technique for spatiotemporal event prediction as it models the rate of event occurrence in terms of both spatial and time points. It is defined as:
\begin{align}
\label{eq:stpp}
    \lambda(t,l|X)=\lim\nolimits_{|\mathrm{d}t|\rightarrow 0,|\mathrm{d}l|\rightarrow 0}\mathbb{E}[N(\mathrm{d}t\times \mathrm{d}l)|X]/(|\mathrm{d}t||\mathrm{d}l|)
\end{align}
Various models have been proposed to instantiate the model of the framework illustrated in Equation \eqref{eq:stpp}. For example, Liu and Brown et al.~\cite{liu2004new} began by assuming there to be a conditional independence among spatial and temporal factors and hence achieved the following decomposition:
\begin{align}
    \lambda(t,l|X)=\lambda(t,l|L,T,F)=\lambda_1(l|L,T,F,t)\cdot\lambda_2(t|T)
\end{align}
where $X,L,T,$ and $F$ denotes the whole input indicator data as well as its different facets, including location, time, and other semantic features, respectively. Then the term $\lambda_1(\cdot)$ can be modeled based on the Markov spatial point process while $\lambda_2(\cdot)$ can be characterized using temporal autoregressive models. To handle situations where explicit assumptions for model distributions are difficult, several methods have been proposed to involve the deep architecture during the point process. Most recently, Okawa et al.~\cite{okawa2019deep} have proposed the following:
\begin{align}
    \lambda(t,l|X)=\int g_\theta\left(t',l',\mathcal{F}(t',l')\right)\cdot\mathcal{K}((t,l),(t',l'))\ \mathrm{d}t'\mathrm{d}l'
\end{align}
where $\mathcal K(\cdot,\cdot)$ is a kernel function such as a Gaussian kernel~\cite{bishop2006pattern} that measures the similarity in time and location dimensions. $\mathcal{F}(t',l')\subseteq F$ denotes the feature values (e.g., event semantics) for the data at location $l'$ and time $t'$. $g_\theta(\cdot)$ can be a deep neural network that is parameterized by $\theta$ and returns an nonnegative scalar. The model selection of $g_\theta(\cdot)$ depends on the specific data types. For example, these authors constructed an image attention network by combining a CNN with the spatial attention model proposed by Lu et al.~\cite{lu2017knowing}.

\subsubsection{Time, location, and Semantics}
\label{sec:time_location_semantics}In this section, we introduce the strategies that jointly predict the time, location, and semantics of future events, which can be grouped into either system-based or model-based strategies.

\noindent\textbf{System-based}. The first type of the system-based methods considered here is the model-fusion system. The most intuitive approach is to leverage and integrate the aforementioned techniques for time, location, and semantics prediction into an event prediction system. For example, a system named EMBERS~\cite{ramakrishnan2014beating} is an online warning system for future events that can jointly predict the time, location, and semantics including the type and population of future events. This system also provides information on the confidence of the predictions obtained. Using an ensemble of predictive models for time~\cite{oki2018mobile}, location, and semantic prediction, this system achieves a significant performance boost in terms of both precision and recall. The trick here is to first prioritize the precision of each individual prediction model by suppressing their recall. Then, due to the diversity and complementary nature of the different models, the fusion of the predictions from different models will eventually result in a high recall. A Bayesian fusion-based strategy has also been investigated~\cite{hoegh2015bayesian}. Another system named \emph{Carbon}~\cite{kang2017carbon} also leverages a similar strategy.

The second type of model involves crowd-sourced systems that implement fusion strategies to generate the event predictions made by the human predictors. For example, in order to handle the heterogeneity and diversity of the human predictors' skill sets and background knowledge under limited human resources, Rostami et al.~\cite{rostami2018crowdsourcing} proposed a recommender system for matching event forecasting tasks to human predictors with suitable skills in order to maximize the accuracy of their fused predictions. Li et al.~\cite{li2016wisdom} took a different approach, designing a prediction market system that operates like a futures market, integrating information from different human predictors to forecast future events. In this system, the predictors can decide whether to buy or sell the ``tokens'' (using virtual dollars, for example) for each specific prediction they have made according to their confidence in it. They typically make careful decisions as they will obtain corresponding awards (for correct predictions) or penalties (for erroneous predictions).

\begin{figure}[htb]
  \centering
    \includegraphics[width=0.9\textwidth]{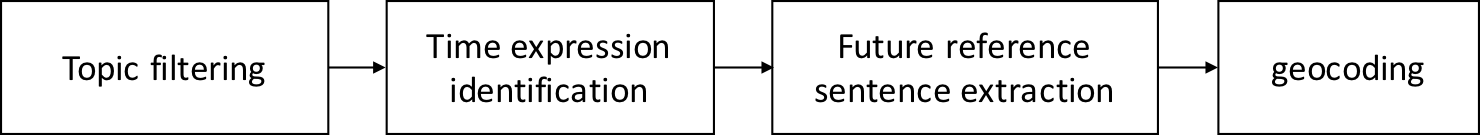}\vspace{-0.3cm}
\caption{Generic framework for planned event detection.\vspace{-0.3cm}}
\label{fig:planned_event}
\end{figure}

\noindent\textbf{Planned future event detection methods}. These methods focus on detecting the planned future events, usually from various media such sources as social media and news and typically relying on NLP techniques and linguistic principles. Existing methods typically follow a workflow similar to the one shown in Figure \ref{fig:planned_event}, consisting of four main steps: 1) \emph{Content filtering.} Methods for content filtering are typically leveraged to retain only the texts that are relevant to the topic of interest. Existing works utilize either supervised methods (e.g., textual classifiers~\cite{kunneman2012leveraging} or unsupervised methods (e.g., querying techniques~\cite{zhao2014unsupervised,muthiah2016capturing}); 2) \emph{Time expression identification} is then utilized to identify future reference expressions and determine the \emph{time to event}. These methods either leverage existing tools such as the Rosetta text analyzer~\cite{dale2018text} or propose dedicated strategies based on linguistic rules~\cite{hurriyetoglu2018estimating}; 3) \emph{Future reference sentence extraction} is the core of planned event detection, and is implemented either by designing regular expression-based rules~\cite{nakajima2017prototype} or by textual classification~\cite{kunneman2012leveraging}; and 4) \emph{Location identification}. The expression of locations is typically highly heterogeneous and noisy. Existing works have relied heavily on geocoding techniques that can resolve the event location accurately. In order to infer the event locations, various types of locations are considered by different researchers, such as article locations~\cite{muthiah2016capturing}, authors' profile locations~\cite{compton2014using}, locations mentioned in the articles~\cite{becker2012identifying}, and authors' neighbors' locations~\cite{jurgens2013s}. Multiple locations have been selected using a geometric median~\cite{compton2014using} or fused using logical rules such as probabilistic soft logic~\cite{muthiah2016capturing}.

\noindent\textbf{Tensor-based methods. }Some methods formulate the data into tensor-form, with dimensions including location, time, and semantics. Tensor decomposition is then applied to approximate the original tensors as the product of multiple low-rank matrices, each of which is a mapping from latent topics to each dimension. Finally, the tensor is extrapolated towards future time periods by various strategies. For example, Mirtaheri~\cite{mirtaheritensor2019tensor} extrapolated the time dimension-matrix only, which they then multiplied with the other dimensions' matrices to recover the estimated extrapolated tensor into the future. Zhou et al.~\cite{zhou2019tensor} took a different approach, choosing instead to add  ``empty values'' for the entries in future time to the original tensor, and then use tensor completion techniques to infer the missing values corresponding to future events.
 \section{Applications of Event Predictions}
\label{sec:applications}
\subsection{Healthcare} This category generally consists of two types of event prediction: 1) population level, which includes disease epidemics and outbreaks, and 2) individual level, which relates to clinical longitudinal events.

\subsubsection{Population-level. }There has been extensive research on disease outbreaks for many different types of diseases and epidemics, including seasonal flu~\cite{adhikari2019epideep}, Zika~\cite{teng2017dynamic}, H1N1~\cite{nishiura2011real}, Ebola~\cite{asher2018forecasting}, and COVID-19~\cite{petropoulos2020forecasting}. These predictions target both the location and time of future events, while the disease type is usually fixed to a specific type for each model. Compartmental models such as SIR models are among the classical mathematical tools used to analyze, model, and simulate the epidemic dynamics~\cite{shaman2012forecasting,zhang2017forecasting}. More recently, individual-based computational models have begun to be used to perform network-based epidemiology based on network science and graph-theoretical models, where an epidemic is modeled as a stochastic propagation over an explicit interaction network among people~\cite{craft2011disease}. Thanks to the availability of high-performance computing resources, another option is to construct a ``digital twin'' of the real world, by considering a realistic representation of a population, including members’ demographic, geographic, behavioral, and social contextual information, and then using individual-based simulations to study the spread of epidemics within each network~\cite{bisset2009epifast}. The above techniques heavily rely on the model assumptions regarding how the disease progresses individually and is transmitted from person to person~\cite{bisset2009epifast}. The rapid growth of large surveillance data and social media data sets such as Twitter and Google flu trends in recent years has led to a massive increase of interest in using   data-driven approaches to directly learn the predictive mapping~\cite{adhikari2019epideep}. These methods are usually both more time-efficient and less dependent on assumptions, while the aforementioned computational models are more powerful for longer-term prediction due to their ability to take into account the specific disease mechanisms~\cite{zhao2015simnest}. Finally, there have also been reports of synergistic research that combines both techniques to benefit from their complementary strengths~\cite{zhao2015simnest,hua2018social}.

\subsubsection{Individual-level. }This research thread focuses on the longitudinal predictive analysis of individual health-related events, including death occurrence~\cite{dempsey2017isurvive}, adverse drug events~\cite{santiso2014adverse}, sudden illnesses such as strokes~\cite{letham2015interpretable} and cardiovascular events~\cite{beunza2019comparison}, as well as other clinical events~\cite{dempsey2017isurvive} and life events~\cite{de2020leveraging} for different groups of people, including elders and people with mental disease. The goal here is usually to predict the time before an event occurs, although some researchers have attempted to predict the type of event. The data sources are essentially the electronic health records of individual patients~\cite{santiso2014adverse,rebane2019investigation}. Recently, social media, forum, and mobile data has also been utilized for predicting drug adverse events~\cite{santiso2014adverse} and events that arise during chronic disease (e.g., chemical radiation and surgery)~\cite{dempsey2017isurvive}.

\subsection{Media} This category focuses on predicting events based on information held in various types of media including: video-based, audio-based, and text-based formats. The core issue is to retrieve key information related to future events utilizing semantic pattern recognition from the data.
\subsubsection{Video- and Audio- based}. While event detection has been extensively researched for video data~\cite{li2017time} and audio mining~\cite{soltau2017neural}, event prediction is more challenging and has been attracting increasing attention in recent years. The goal here is usually to predict the future status of the objects in the video, such as the next action of soccer players~\cite{decroos2017predicting} or basketball players~\cite{Neumann_2019_CVPR_Workshops}, or the movement of vehicles~\cite{yuen2010data}. 

\subsubsection{Text- and script- based} A huge amount of news data has accumulated in recent decades, much of which can be used for big data predictive analytics among news events. A number of researchers have focused on predicting the location, time, and semantics of various events. To achieve this, they usually leverage the immense historical news and knowledge base in order to learn the association and causality among events, which is then applied to forecast events when given current events. Some studies have even directly generated textual descriptions of future events by leveraging NLP techniques such as sequence to sequence learning~\cite{radinsky2012learning,radinsky2013mining,hu2017happens,xue2018miml,lei2019event,su2020hierarchical,dami2018news,shrestha2017predicting,nakajima2017prototype}.

\subsection{Transportation}. This category can be classified into: 1) population based events, including dispersal events, gathering events, and congestion; and 2) individual-level events, which focus on fine-grained patterns such as human mobility behavior prediction.

\subsubsection{Group transportation pattern}. Here, researchers typically focus on transportation events such as congestion~\cite{chen2018pcnn,jiang2019deepurbanevent}, large gatherings~\cite{vahedian2017forecasting}, and dispersal events~\cite{vahedian2019predicting}. The goal is thus to forecast the future time period~\cite{fu2019titan} and location~\cite{vahedian2017forecasting} of such events. Data from traffic meters, GPS, and mobile devices are usually used to sense real-time human mobility patterns. Transportation and geographical theories are usually considered to determine the spatial and temporal dependencies for predicting these events.

\subsubsection{Individual transportation behavior} Another research thread focuses on individual-level prediction, such as predicting an individual's next location~\cite{li2018next,yang2018spatio} or the likelihood or time duration of car accidents~\cite{bao2019spatiotemporal,ren2018deep,yuan2018hetero}. Sequential and trajectory analyses are usually used to process trajectory and traffic flow data.

\subsection{Engineering Systems} Different types of engineering systems have begun to routinely apply event forecasting methods, including: 1) civil engineering, 2) electrical engineering, 3) energy engineering, and 4) other engineering domains. Despite the variety of systems in these widely different domains, the goal is essentially to predict future abnormal or failure events in order to support the system's sustainability and robustness. Both the location and time of future events are key factors for these predictions. The input features usually consist of sensing data relevant to specific engineering systems.
\begin{itemize}
\item\textbf{Civil engineering.} This covers various a wide range of problems in diverse urban systems, such as smart building fault adverse event prediction~\cite{basak2016scalable}, emergency management equipment failure prediction~\cite{ding2018online}, manhole event prediction~\cite{rudin2010process}, and other events~\cite{huang2019mist}.
\item\textbf{Electrical engineering}. This includes teleservice systems failures~\cite{deep2019event} and unexpected events in wire electrical discharge machining operations~\cite{sanchez2018unexpected}. 
\item\textbf{Energy engineering}. Event prediction is also a hot topic in energy engineering, as such systems usually require strong robustness to handle the disturbance from the nature environments. Active research domains here include wind power ramp prediction~\cite{gallego2015review}, solar power ramp prediction~\cite{abuella2019forecasting}, and adverse events in low carbon energy production~\cite{coniglio2020infrequent}.
\item\textbf{Other engineering domains}. There is also active research on event prediction in other domains, such as irrigation event prediction  in agricultural engineering~\cite{perea2019prediction} and mechanical fault prediction in mechanical engineering~\cite{susto2014machine}.
\end{itemize}
\subsection{Cyber Systems} Here, the prediction models proposed generally focus on either network-level events or device-level events. For both types, the general goal is essentially to predict the likelihood of future system failure or attacks based on various indicators of system vulnerability. So far these two categories have essentially differed only in their inputs: the former relies on network features, including system specifications, web access logs and search queries, mismanagement symptoms, spam, phishing, and scamming activity, although some researchers are investigating the use of social media text streams to identify semantics indicating future potential attack targets of DDoS~\cite{matsubara2012fast,wang2017ddos}. For device-level events, the features of interest are usually the binary file appearance logs of machines~\cite{oki2018mobile,wang2015iot}. Some work has been done on micro-architectural attacks~\cite{gulmezoglu2019fortuneteller} by observing and proactively analyzing the observations on speculative branches, out-of-order executions and shared last level caches~\cite{shen2018tiresias}.




\subsection{Political events}
Political event prediction has become a very active research area in recent years, largely thanks to the popularity of social media. The most common research topics can be categorized as: 1) offline events, and 2) online activism. 

\subsubsection{Offline events} This includes civil unrest~\cite{ramakrishnan2014beating}, conflicts~\cite{ward2010perils}, violence~\cite{blair2017predicting}, and riots~\cite{dos2014forecasting}. This type of research usually targets the future events' geo-location, time, and topics by leveraging the social sensors that indicate public opinions and intentions. Utilization of social media has become a popular approach for these endeavors, as social media is a source of vital information during the event development stage~\cite{ramakrishnan2014beating}. Specifically, many aspects are clearly visible in social media, including complaints from the public (e.g., toward the government), discussions about their intentions regarding specific political events and targets, as well as advertisements for the planned events. Due to the richness of this information, further information on future events such as the type of event~\cite{gao2019incomplete}, the anticipated participants population~\cite{ramakrishnan2014beating}, and the event scale~\cite{gao2018incomplete} can also be discovered in advance.

\subsubsection{Online events} Due to the major impact of online media such as online forums and social media, many events such as online activism, petitions, and hoaxes in such online platform also involve strong motivations for achieving some political purpose~\cite{wang2018incomplete}. Beyond simple detection, the prediction of various types of events have been studied in order to enable proactive intervention to sidetrack the events such as hoaxes and rumor propagation~\cite{jin2013epidemiological}. Other researchers have sought to foresee the results of future political events in order to benefit a particular group of practitioners, for example by predicting the outcome of online petitions or presidential elections~\cite{wang2018incomplete}.

\subsection{Natural disasters}
Different types of natural disasters have been the focus of a great deal of research. Typically, these are rare events, but mechanistic models, a long historical records (often extending back dozens or hundreds of years), and domain knowledge are usually available. The input data are typically collected by sensors or sensor networks and the output is the risk or hazard of future potential events. Since these event occurrence are typically rare but very high-stakes, many researchers strive to cover all event occurrences and hence aim to ensure high recalls. 

\subsubsection{Geophysics-related}
\textbf{Earthquakes.} Predictions here typically focus on whether there will be an earthquake with a magnitude larger than a specified threshold in a certain area  during a future period of time. To achieve this, the original sensor data is usually proccessed using geophysical models such as Gutenberg–Richter’s inverse law, the distribution of characteristic earthquake magnitudes, and seismic quiescence~\cite{asim2017earthquake,reyes2013neural}. The processed data are then input into machine learning models that treat them as input features for predicting the output, which can be either binary values of event occurrence or time-to-event values. Some studies are devoted to identifying the time of future earthquakes and their precursors, based on an ensemble of regressors and feature selection techniques~\cite{rouet2017machine}, while others focus on aftershock prediction and the consequences of the earthquake, such as fire prevention~\cite{mallouhy2019major}. It worth noting that social media data has also been used for such tasks, as this often supports early detection of the first-wave earthquake, which can then be used to predict the afterstocks or earthquakes in other locations~\cite{sakaki2010earthquake}.

\textbf{Fire events}. Research in this category can be grouped into urban fires and wildfires. This type of research often focuses on the time at which a fire will affect a specific location, such as a building. The goal here is to predict the risk of future fire events. To achieve this, both the external environment and the intrinsic properties of the location of interests are important. Therefore, both static input data (e.g., natural conditions and demographics) and time-varying data (e.g., weather, climate, and crowd flow) are usually involved. Shin and Kim~\cite{shin2019autoencoder} focus on building fire risk prediction, where the input is the building's profile. Others have studied wildfires, where weather data and satellite data are important inputs. This type of research focuses primarily on predicting both the time and location of future fires~\cite{YUCHI2016308,wang2019cityguard}.

Other researchers have focused on rarer events such as volcanic eruptions. For example, some leverage chemical prior knowledge to build a Bayesian network for prediction ~\cite{cheon2009bayesian}, while others adopt point processes to predict the hazard of future events~\cite{damaschke2018volcanic}.

\subsubsection{Atmospheric science-related}

\textbf{Flood events}. Floods may be caused by many different reasons, including atmospheric (e.g., snow and rain), hydrological (e.g., ice melting, wind-generated waves, and river flow), and geophysical (e.g., terrain) conditions. This makes the forecasting of floods highly complicated task that requires multiple diverse predictors~\cite{wang2013towards}. Flood event prediction has a long history, with the latest research focusing especially on computational and simulation models based on domain knowledge. This usually involves using ensemble prediction systems as inputs for hydrological and/or hydraulic models to produce river discharge predictions. For a detailed survey on flood computational models please refer to~\cite{cloke2009ensemble}. However, it is prohibitively difficult to comprehensively consider and model all the factors correctly while avoiding all the accumulated errors from upstream predictions (e.g., precipitation prediction). Another direction, based on data-driven models such as statistical and machine learning models for flood prediction, is deemed promising and is expected to be complementary to existing computational models. These newly developed machine learning models are often based solely on historical data, requiring no knowledge of the underlying physical processes. Representative models are SVM, random forests, and neural networks and their variants and hybrids. A detailed recent survey is provided in~\cite{mosavi2018flood}.

\textbf{Tornado Forecasting}. Tornadoes usually develop within thunderstorms and hence most tornado warning systems are based on the prediction of thunderstorms. For a comprehensive survey, please refer to~\cite{doswell1993tornado}. Machine learning models, when applied to tornado forecasting tasks, usually suffer from high-dimensionality issues, which are very common in meteorological data. Some methods have leveraged dimensional reduction strategies to preprocess the data~\cite{yu2015tornado} before prediction. Research on other atmosphere-related events such as droughts and ozone events has also been conducted~\cite{field2012managing}.

\subsubsection{Astrophysics-related} There is a large body of prediction research focusing on events outside the Earth, especially those affecting the star closest to us, the sun. Methods have been proposed to predict various solar events that could  impact life on Earth, including solar flares~\cite{barnes2016comparison}, solar eruptions~\cite{aggarwal2018prediction}, and high energy particle storms~\cite{martens2017data}. The goal here is typically to use satellite imagery data of the sun to predict the time and location of future solar events and the activity strength~\cite{falconer2014mag4}.


\subsection{Business}
Business intelligence can be grouped into company-based events and customer-based events. 

\subsubsection{Customer activity prediction} The most important customer activities in business is whether a customer will continue doing business with a company and how long a costumer will be willing to wait before receiving the service? A great deal of research has been devoted to these topics, which can be categorized based on the type of business entities namely enterprises, social media, and education, who are primarily interested in churn prediction, site migration, and student dropout, respectively. The first of these focuses on predicting whether and when a customer is likely to stop doing business with a profitable enterprise~\cite{eria2018systematic}. The second aims to predict whether a social media user will move from one site, such as Flickr, to another, such as Instagram, a movement known as site migration~\cite{zafarani2014social}. While site migration is not popular, attention migration might actually be much more common, as a user may ``move'' their major activities from one social media site to another. The third type, student dropout, is a critical domain for education data mining, where the goal is to predict the occurrence of absenteeism from school for no good reason for a continuous number of days; a comprehensive survey is available in~\cite{mduma2019survey}. For all three types, the procedure is first to collect features of a customer's profile and activities over a period of time and then conventional or sequential classifiers or regressors are generally used to predict the occurrence or time-to-event of the future targeted activity.








\subsubsection{Business process events} Financial event prediction has been attracting a huge amount of attention for risk management, marketing, investment prediction and fraud prevention. Multiple information resources, including news, company announcements, and social media data could be utilized as the input, often taking the form of time series or temporal sequences. These sequential inputs are used for the prediction of the time and occurrence of future high-stack events such as company distress, suspension, mergers, dividends, layoffs, bankruptcy, and market trends (rises and falls in the company's stock price)~\cite{yang2019using,mehdiyev2017multi,cecchini2010making,le2012hybrid,chan2011text,hagenau2012automated,ding2015deep}.

\subsection{Crime}
It is difficult to deduce the precise location and time for individual crime incidences. Therefore, the focus is instead estimating the risk and probability of the location, time, and types of future crimes. This field can be naturally categorized based on the various crime types:
\subsubsection{Political crimes and terrorism}
This type of crime is typically highly destructive, and hence attracts huge attention in its anticipation and prevention. Terrorist activities are usually aimed at religious, political, iconic, economic or social targets. The attacker typically targets larger numbers of people and the evidences related to such attacks is retained in the long run. Though it is extremely challenging to predict the precise location and time of individual terrorism incidents, numerous studies have shown the potential utility for predicting the regional risks of terrorism attacks based on information gathered from many data sources such as geopolitical data, weather data, and economics data. The Global Terrorism Database is the most widely recognized dataset that records the descriptions of world-wide terrorism events of recent decades. In addition to terrorism events, other similar events such as mass killings~\cite{ulfelder2013multimodel} and armed-conflict events~\cite{spagat2018fundamental} have also been studied using similar problem formulations.


\subsubsection{Crime incidents} Most studies on this topic focus on predicting the types, intensity, count, and probability of crime events across defined geo-spatial regions. Until now, urban crimes are most commonly the topic of research due to data availability. The geospatial characteristics of the urban areas, their demographics, and temporal data such as news, weather, economics, and social media data are usually used as inputs. The geospatial dependency and correlation of the crime patterns are usually leveraged during the prediction process using techniques originally developed for spatial predictions, such as kernel density estimation and conditional random fields. Some works simplify the tasks by only focusing on specific types of crimes such as theft~\cite{rumi2018theft}, robbery, and burglary~\cite{cortez2018architecture}.

\subsubsection{Organized and serial crimes}

Unlike the above research on regional crime risks, some recent studies strive to predict the next incidents of criminal individuals or groups. This is because different offenders may demonstrate different behavioral patterns, such as targeting specific regions (e.g., wealthy neighborhoods),  victims (e.g., women), for specific benefits (e,g, money). The goal here is thus is to predict the next crime site and/or time, based on the historical crime event sequence of the targeted criminal individual or group. Models such as point processes~\cite{li2018next} or Bayesian networks~\cite{liao2010novel} are usually used to address such problems.
 \section{Open Challenges and Future Directions}
\label{sec:discussion}
Despite the major advances in event prediction in recent years, there are still a number of open problems and potentially fruitful directions for future research, as follows:

\subsection{Model Transparency, Interpretability, and Accountability} Increasingly sophisticated forecasting models have been proposed to improve the prediction accuracy, including those utilizing approaches such as ensemble models, neural networks, and the other complex systems mentioned above. However, although the accuracy can be improved, the event prediction models are rapidly becoming too complex to be interpreted by human operators. The need for better model accountability and interpretability is becoming an important issue; as big data and Artificial Intelligence techniques are applied to ever more domains this can lead to serious consequences for applications such as healthcare and disaster management. Models that are not interpretable by humans will find it hard to build the trust needed if they are to be fully integrated into the workflow of practitioners. A closely related key feature is the accountability of the event prediction system. For example, disaster managers need to thoroughly understand a model's recommendations if they are to be able to explain the reason for a decision to displace people in a court of law. Moreover, an ever increasing number of laws in countries around the world are beginning to require adequate explanations of decisions reached based on model recommendations. For example, Articles 13-15 in the European Union's General Data Protection Regulation (GDPR)~\cite{voigt2017eu} require algorithms that make decisions that “significantly affect" individuals to provide explanations (``right to explanation'') by May 28, 2018. Similar laws have also been established in countries such as the United States~\cite{coglianese2016regulating} and China~\cite{qi2018assessing}.

\subsection{Vulnerability to Noise and Adversarial Attacks} 

The massive popularity of the proposal, development, and deployment of event prediction is stimulating a surge interest in developing ways to counter-attack these systems. It will therefore not be a surprise when we begin to see the introduction of techniques to obfuscate these event prediction methods in the near future. As with many state-of-the-art AI techniques applied in other domains such as object recognition, event prediction methods can also be very vulnerable to noise and adversarial attacks. The famous failure of Google Flu trends, which missed the peak of the 2013 flu season by 140 percent due to low relevance and high disturbance affecting the input signal, is a vivid memory for practitioners in the field~\cite{fung2014google}. Many predictions relying on social media data can also be easily influenced or flipped by injecting scam messages. Event prediction models also tend to over-rely on low-quality input data that can be easily disturbed or manipulated, lacking sufficient robustness to survive noisy signals and adversarial attacks. Similar problems threaten to other application domains such as business intelligence, crime, and cyber systems.

\subsection{Integration of mechanistic knowledge and data driven-models} Over the years, many domains have accumulated a significant amount of knowledge and experience about event development occurrence mechanisms, which can thus provide important clues for anticipating future events, such as epidiomiology models, socio-political models, and earthquake models. All of these models focus on simplifying real-world phenomena into concise principles in order to grasp the core mechanism, discarding many details in the process. In contrast, data-driven models strive to ensure the accurate fitting of large historical data sets, based on sufficient model expressiveness but cannot guarantee that the true underlying principle and causality of event occurrence modeled accurately. There is thus a clear motivation to combine their complementary strengths, and although this has already attracted great deal of interest~\cite{zhao2015simnest,hua2018social}, most of the models proposed so far are merely ensemble learning-based and simply merge the final predictions from each model. A more thorough integration is needed that can directly embed the core principles to regularize and instruct the training of data-driven event prediction methods. Moreover, existing attempts are typically specific to particular domains and are thus difficult to develop further as they require in-depth collaborations between data scientists and domain experts. A generic framework developed to encompass multiple different domains is imperative and would be highly beneficial for the various domain experts.

\subsection{Prescriptive and counterfactual analysis} The ultimate purpose of event prediction is usually not just to anticipate the future, but to change it, for example by avoiding a system failure and flattening the curve of a disease outbreak. However, it is difficult for practitioners to determine how to act appropriately and implement effective policies  in order to achieve the desired results in the future. This requires a capability that goes beyond simply predicting future events based on the current situation, requiring them instead to also take into account the new actions being taken in real time and then predict how they might influence the future. One promising direction is the use of counterfactual event~\cite{menzies2001counterfactual} prediction that models what would have happened if different circumstances had occurred. Another related direction is prescriptive analysis where different actions can be merged into the prediction system and future results anticipated or optimized. Related works have been developed in few domains such as epidemiology. However, as yet these lack sufficient research in many other domains that will be needed if we are to develop generic frameworks that can benefit different domains. 


\subsection{Multi-objective training} Existing event prediction methods mostly focus primarily on accuracy. However, decision makers who utilize these predicted event results usually need much more, including key factors such as event resolution (e.g., time resolution, location resolution, description details), confidence (e.g., the probability a predicted event will occur), efficiency (whether the model can predict per day or per seccond), lead time (how many days the prediction can be made prior to the event occurring), and event intensity (how serious it is). multi-objective optimization (e.g., accuracy, confidence, resolution). There are typically trade-offs among all the above metrics and accuracy, so merely optimizing accuracy during training will inevitably mean the results drift away from the overall optimal event-prediction-based decision. A system that can flexibly balance the trade-off between these metrics based on decision makers' needs and achieve a multi-objective optimization is the ultimate objective for these models.

\section{Conclusion}
\label{sec:conclusions}
This survey has presented a comprehensive survey of existing methodologies developed for event prediction methods in the big data era. It provides an extensive overview of the event prediction challenges, techniques, applications, evaluation procedures, and future outlook, summarizing the research presented in over 200 publications, most of which were published in the last five years. Event prediction challenges, opportunities, and formulations have been discussed in terms of the event element to be predicted, including the event location, time, and semantics, after which we went on to propose a systematic taxonomy of the existing event prediction techniques according to the formulated problems and types of methodologies designed for the corresponding problems. We have also analyzed the relationships, differences, advantages, and disadvantages of these techniques from various domains, including machine learning, data mining, pattern recognition, natural language processing, information retrieval, statistics, and other computational models. In addition, a comprehensive and hierarchical categorization of popular event prediction applications has been provided that covers domains ranging from natural science to the social sciences. Based upon the numerous historical and state-of-the-art works discussed in this survey, the paper concludes by discussing open problems and future trends in this fast-growing domain.

\bibliographystyle{ACM-Reference-Format}
 \bibliography{acmart}
\end{document}